\documentclass{article}
\usepackage{latexsym}
\usepackage{amsfonts}
\usepackage{graphicx}

\newtheorem{theorem}{Theorem}[section]
\newtheorem{lemma}{Lemma}[section]
\newtheorem{cor}{Corollary}[section]
\newtheorem{definition}{Definition}[section]
\newtheorem{prop}{Proposition}[section]
\newcommand{\qed}{\hfill$\Box$\par\medskip}

\newenvironment{Proof}{\noindent{\sc Proof.}}{\qed}

\def\bhag#1{\noindent
\setcounter{equation}{0}
\section{#1}
}
\def\bfgk#1{{{#1}\kern-5.5pt{#1}}}

\def\RR{{\mathbb R}}
\def\CC{{\mathbb C}}
\def\XX{{\mathbb X}}
\def\GG{{\mathbb G}}

\def\PPI{{{\rm I}\kern-1pt\Pi}}

\def\G{{\mathcal G}}
\def\V{{\mathcal V}}
\def\GG{{\mathbb G}}

\def\a{\alpha}
\def\b #1;{{\bf #1}}

\def\y{{\bf y}}

\def\e{\epsilon}

\def\O{{\cal O}}
\def\P{{\cal P}}

\def\C{{\mathcal C}}
\def\W{{\bf W}}

\def\ip#1#2{{\langle {#1}, {#2}\rangle}}
\def\derf#1#2{{#1}^{(#2)}}

\def\esssup{\mathop{\hbox{{\rm ess sup}}}}

\def\be{\begin{equation}}
\def\ee{\end{equation}}
\def\bea{\begin{eqnarray}}
\def\eea{\end{eqnarray}}
\def\eref#1{(\ref{#1})}
\def\disp{\displaystyle}
\def\dist{\mbox{\textsf{ dist }}}

\def\span{\mbox{{\textsf{span }}}}
\def\donchitre#1#2{\vskip 6.5cm\noindent
\parbox[t]{1in}{\special{eps:#1.eps x=6.5cm y=5.5cm}}
\hbox to 7cm{}\parbox[t]{0.0cm}{\special{eps:#2.eps x=6.5cm y=5.5cm}}}

\begin{document}
%T
\title{Eignets for function approximation on manifolds}
\author{
H.~N.~Mhaskar\thanks{Department of Mathematics, California State University,
Los Angeles, California, 90032, USA,
\textsf{email:} hmhaska@calstatela.edu. The research of this author was supported, in part,
by grants  from the National Science Foundation and  the U.S.  Army Research Office. }
}
\date{}           
\maketitle

%A
\begin{abstract}
Let $\XX$ be a compact, smooth, connected, Riemannian manifold without boundary, $G:\XX\times\XX\to \RR$ be a kernel. Analogous to a radial basis function network, an eignet is an expression of the form $\sum_{j=1}^M a_jG(\circ,y_j)$, where $a_j\in\RR$, $y_j\in\XX$, $1\le j\le M$. We describe a deterministic, universal algorithm for constructing an eignet for approximating functions in $L^p(\mu;\XX)$ for a general class of measures $\mu$ and kernels $G$. Our algorithm yields linear operators. Using the minimal separation amongst the centers $y_j$ as the cost of approximation, we give modulus of smoothness estimates for the degree of approximation by our eignets, and show by means of a converse theorem that these are the best possible for every \emph{individual function}. We also give estimates on the coefficients $a_j$ in terms of the norm of the eignet. Finally, we demonstrate that if any sequence of eignets satisfies the optimal estimates for the degree of approximation of a smooth function, measured in terms of the minimal separation, then the derivatives of the eignets also approximate the corresponding derivatives of the target function in an optimal manner.
\end{abstract}


%1
\bhag{Introduction}
In recent years, diffusion geometry techinques have developed into a powerful tool for analysis of a nominally high dimensional data, which has a low dimensional structure, for example, it lies on a low dimensional manifold in the high dimensional ambient space. Applications of these techniques  include document analysis \cite{mauro1}, face recognition \cite{niyogi}, semi--supervised learning \cite{niyogi1, niyogi2}, image processing \cite{donoho1}, and cataloguing of galaxies \cite{donoho2}. The special issue  \cite{achaspissue} of Applied and Computational Harmonic Analysis contains several papers that serve as a good introduction to this subject.  

An essential ingredient in these techniques is the notion of a heat kernel $K_t$ on the manifold $\XX$ in question, which can be defined formally by
$$
K_t(x,y)=\sum_{j\ge 0}\exp(-\ell_j^2 t)\phi_j(x)\phi_j(y), \qquad t>0,\ x,y\in\XX,
$$
where  $\{\phi_j\}$  is an orthonormal basis for $L^2(\mu;\XX)$ for an appropriate measure $\mu$, and $\ell_j$'s are nonnegative numbers increasing to $\infty$ as $j\to\infty$. A multiresolution analysis is then defined by Coifmann and Maggioni \cite{mauro1} for a fixed $\epsilon>0$ by defining the increasing sequence of scaling spaces 
$$
\mbox{span }\{\phi_k : \exp(-2^{-j}\ell_k^2) \ge \epsilon\}=\mbox{span }\{\phi_k : \ell_k^2\le (2^j\log(1/\epsilon))\}.
$$
 The range of the operators generated by $K_{2^{-j}}$ being ``close'' to the space at level $j$, one may obtain an approximate projection of a function by applying these operators to the function. In turn, these operators can be computed using fast multipole techniques. The diffusion wavelets and wavelet packets can be obtained by applying Gram Schmidt procedure to the kernels $K_{2^{-j}}$. On a more theoretical side, Jones, Maggioni, and Schul \cite{jms} have recently proved that the heat kernel can be used to construct a local coordinate atlas on manifolds, preserving the order of magnitude of the distances between points within each chart. 

Since an explicit formula for the heat kernel is typically not known on all but the simplest of manifolds, in  numerical implementations, one considers in place of the heat kernel an approximation by means of a suitable radial basis function, typically a Gaussian. The error in this approximation is investigated in detail by several authors, for example, \cite{lafon, amit, belkinfound, belkinspectconv}. 
In a different idea, Saito \cite{saito} has advocated the use of other kernels which commute with the heat kernel, and hence, share the invariant subspaces with it, but for which explict formulas are known.

Several applications, especially in the context of semi--supervised learning, signal processing, and pattern recognition can be viewed as problems of function approximation. For example, given a few digitized images of handwritten digits, one wishes to develop a model that will predict for any other image whether the corresponding digit is 0. Each image may be viewed as a point in a high dimensional space, and the target function is the characteristic function of the set of points corresponding to the digit 0. We observe in this context that 
even though ${\mathcal K}_tf\to f$ (uniformly if $f$ is continuous) as $t\to 0$, where ${\mathcal K}_t$ is the heat operator defined by the kernel $K_t$, the rate of convergence provided by this simple minded approximation cannot be the optimal one for smooth functions, since the ${\mathcal K}_t\phi_j\not=\phi_j$ except when $\ell_j=0$. In this paper, for $L>0$, an element of $\Pi_L:=\span\{\phi_j : \ell_j\le L\}$ will be called a \emph{diffusion polynomial} of degree at most $L$, as in \cite{mauropap}. In \cite{fasttour, mauropap}, we have developed a different multiscale analysis based on $\Pi_{2^j}$ as the scaling spaces.  We have obtained a Littlewood--Paley expansion, valid for functions in all $L^p$ spaces \emph{including $p=1,\infty$}. This expansion is in terms of a tight frame transform, which can be used to characterize different Besov spaces related to approximation by diffusion polynomials. Our tight frames can also be chosen to be highly localized. 

 The main objective of this paper is to consider the approximation properties of a generalized translation network  of the form $\sum_{j=1}^M a_jG(\circ,y_j)$, where $G$ is a fixed kernel, $G:\XX\times\XX\to \RR$, $M\ge 1$ is an integer (the number of \emph{neurons}), the coefficients $a_j$'s are real numbers and the \emph{centers} $y_j$'s are distinct points in $\XX$.  We will deal with kernels of the form $G(x,y)=\sum_{j=0}^\infty b(\ell_j)\phi_j(x)\phi_j(y)$. For this reason, we will call the network an \emph{eignet}. This paper is the first part of a two part investigation. In this paper, we consider the case when $\{b(\ell_j)\ell_j^\beta\}$ remains bounded as $j\to\infty$; in a sequel, we plan to develop analogous theory for the case when $\{b(\ell_j)\}$ tends to $0$ exponentially fast as $j\to\infty$, in particular, including the case of the heat kernel itself as $G$. 

To explain our objectives in further detail, we describe first the general paradigm in approximation theory. Typically, one considers a metric space ${\mathcal X}$ and a nested, increasing sequence of subsets of ${\mathcal X}$: $V_0\subset V_1\subset \cdots V_m\subset V_{m+1}\subset \cdots$. Elements of $V_m$ provide a model (\emph{approximant})  for a \emph{target function} $f\in {\mathcal X}$; the index $m$ is typically related to the model complexity.  The \emph{density theorem} is a statement that $\cup_{m=0}^\infty V_m$ is dense in ${\mathcal X}$. Let $d({\mathcal X};f,g)$ denote the distance between $f, g\in {\mathcal X}$. A deeper, and central problem of approximation theory is to investigate the rate at which the \emph{degree of approximation}, $\dist({\mathcal X};f, V_m):=\inf_{P\in V_m}\dist ({\mathcal X};f,P)$, converges to $0$ as $m\to\infty$, depending upon certain conditions on $f$. These conditions are encoded by a statement that $f\in W$ for a subset $W\subset {\mathcal X}$, usually called a \emph{smoothness class}.  In the most classical example, the \emph{trigonometric case}, ${\mathcal X}$ is the space of all continuous, $2\pi$--periodic functions on $\RR$, equipped with the supremum norm on $[-\pi,\pi]$, and $V_m$ denotes the class of all trigonometric polynomials of order at most $m$; i.e., expressions of the form $\sum_{|j|\le m} a_je^{ij\circ}$. The well known \emph{equivalence theorem} in this case states \cite{devlorbk} that if $0<\alpha<1$, and $r\ge 0$ is an integer, then $\dist ({\mathcal X};f,V_m)=\O(m^{-r-\a})$ if and only if $f$ has $r$ continuous derivatives and $|\derf{f}{r}(x)-\derf{f}{r}(y)|=\O(|x-y|^\a)$, $x,y\in\RR$. To cover the case when $\a=1$ is allowed, one needs to introduce higher order moduli of smoothness; a more modern approach is to consider  $K$ functionals. We observe that this theory is applicable to individual functions, rather than being an assertion about the existence of a function to demonstrate that the rate at which the degree of approximation converges to zero cannot be improved.  In the general case, of course, the interesting questions are to determine what one should mean by the model complexity, and what smoothness classes are \emph{characterized} by  a given rate of convergence of $\dist ({\mathcal X};f,V_m)$ to $0$ as $m\to\infty$. In the context of approximation by Gaussian networks, we have demonstrated in \cite{convtheo, mhasbook} that a satisfactory theory can be developed by using the minimal separation amongst the centers as the measurement of model complexity, with the smoothness classes defined in terms of certain weighted Besov spaces. 

The main goal of this paper is to demonstrate equivalence theorems of approximation theory in the case of eignets, where the complexity of the model is measured by the minimal separation amongst the centers and the smoothness of the target function is measured by a suitable $K$ functional as in \cite{mauropap}.  In this paper, we will show that the smoothness classes characterized by the degrees of approximation by eignets with minimal separation $q$ amongst the centers  are the same as those characterized by the degrees of approximation by $\Pi_{1/q}$, $q\to 0$. 

There are several consequences of our approach, which we find interesting. First, we will give an explicit, stable, construction of an eignet, which is universal in the sense that it is defined for every function in $L^p$ (or every continuous function, depending upon the data available for the function). At the same time, the approximation error  for any individual function in a smoothness class is commensurate with the degree of approximation by the class of all eignets with the same minimal separation amongst the centers. Our operator will automatically minimize (up to a constant multiple) a regularization criterion, but does not require the solution of an optimization problem to achieve this.

Second, for an arbitrary eignet, we will estimate the size of the coefficients in terms of the norm of the eignet itself. This estimate will be in terms of the minimal separation amongst the centers. In particular, if one wishes to interpolate using eignets, our result gives an estimate on the stability of the interpolation matrix. Finally, we will consider the question of simultaneous approximation: if $\Psi$ is an arbitrary eignet, and one knows an upper bound for $\|f-\Psi\|_p$, we estimate the error $\|(\Delta^*)^r f-(\Delta^*)^r\Psi\|_p$, where $\Delta^*$ is a pseudo--differential operator. 

One of the referees has pointed out kindly that our work here has several potential applications: signal processing, Paley Wiener theorems in inverse problems, computer vision, imaging, geo-remote sensing, among others, and that further hints can be found in \cite{damelin1, damelin2, damelin3, damelin4, damelin5}. 

The paper is organized as follows. In Section~\ref{setupsect}, we will describe the general set up, including the conditions on the manifold, the system $\{\phi_j\}$, the kernel $G$, etc., including some basic facts. The main results are described in Section~\ref{mainsect}. The proofs of these results involve a great deal of estimations involving many sums and integrals. These estimations being very similar, we prefer to present them concisely in a somewhat abstract setting. This setting and the appearance which the various objects  in Section~\ref{mainsect} take is explained in Section~\ref{abstractsect}. Several preparatory lemmas and propositions  of a technical nature are proved in Section~\ref{techsect}. In Section~\ref{proofsect}, we use these  to prove the new results in Section~\ref{mainsect}. In a first reading, one may wish to skip Section~\ref{techsect}, and refer back to it as needed from Section~\ref{proofsect}.

We thank the referees and the editor for their many valuable suggestions for the improvement of the first draft of this paper.
We thank J\"urgen Prestin and Frank Filbir for their encouragement and discussions during the preparation of this paper.

%2
\bhag{The set up}\label{setupsect}
Our results in this paper involve a number of objects: the Riemannian manifold $\XX$, the geodesic distance $\rho$ on $\XX$, a measure $\mu$ on $\XX$, the system $\{\phi_j\}$, the sequence $\{\ell_j\}$, the kernel $G$ for the eignet, etc. In this section, we introduce the notations and various assumptions on these objects.

\subsection{The manifold}\label{manifoldsect}
Throughout this paper, $\XX$ is assumed to be a ($C^\infty$) smooth, compact, connected, Riemannian manifold, $\rho$ denotes the geodesic distance on $\XX$, $\mu$ is a fixed probability measure on $\XX$, not necessarily the manifold measure on $\XX$. For $x\in\XX$, $r>0$, let
$$
B(x.r):=\{y\in\XX\ :\ \rho(x,y)\le r\}, \ \Delta(x,r)=\XX\setminus B(x,r).
$$
We assume that there exists $\a>0$ such that
\be\label{ballmeasurecond}
\mu(B(x,r))\le cr^\a, \qquad x\in\XX, \ r>0.
\ee
Here, and in the sequel, the symbols $c, c_1,\cdots$ will denote generic positive constants depending only on the fixed parameters in the discussion, such as $\rho$, $\mu$,  the system $\{\phi_k\}$,  and the norms, etc. Their value may be different at different occurrences, even within a single formula. The notation $A\sim B$ means that $c_1A\le B\le c_2A$. 

If $X\subseteq \XX$ is $\mu$-measurable, and $f : X\to \CC$ is a $\mu$-measurable function, we will write
$$
\|f\|_{X,p}:=\left\{\begin{array}{ll}
\disp \left\{\int_{X}|f(x)|^pd\mu(x)\right\}^{1/p}, & \mbox{ if $1\le p<\infty$,}\\
\mu-\esssup_{x\in X}|f(x)|, &\mbox{ if $p=\infty$.}
\end{array}\right.
$$
The class of all $f$ with $\|f\|_{X,p}<\infty$ will be denoted by $L^p(X)$, with the usual convention of considering two functions to be equal if they are equal $\mu$--almost everywhere. If $X=\XX$, we will omit its mention from the notations. For $1\le p\le\infty$, we define $p'=p/(p-1)$ with the usual understanding that $1'=\infty$, $\infty'=1$. If $f_1\in L^p$, $f_2\in L^{p'}$ then
$$
\ip{f_1}{f_2}:=\int_\XX f_1(x)f_2(x)d\mu(x).
$$
If $f\in L^p$, $W\subseteq L^p$, we define
$$
\dist(p;f,W):=\inf_{P\in W}\|f-P\|_p,
$$
an abbreviation for $\dist(L^p;f,W)$.

Let $\{\phi_j\}$ be an orthonormal system of functions in $L^2$, such that each $\phi_j$ is continuous on $\XX$ (and hence, both integrable and bounded). We assume that $\phi_0(x)\equiv 1$ for $x\in\XX$. Let $\{\ell_j\}$ be a nondecreasing sequence of real numbers such that $\ell_0=0$, $\ell_j\uparrow \infty$ as $j\to\infty$. For $L\ge 0$, we write $\Pi_L:=\span\{\phi_j : \ell_j\le L\}$. An element of $\Pi_\infty:=\cup_{L\ge 0}\Pi_L$ will be called a diffusion polynomial. For $P\in\Pi_\infty$, the degree of $P$ is the minimum integer $L$ such that $P\in\Pi_L$. The $L^p$ closure of $\Pi_\infty$ will be denoted by $X^p$.

 For $t>0$, $x,y\in\XX$, we define the \emph{heat kernel} on $\XX$ formally by
\be\label{heatkerndef}
K_t(x,y)=\sum_{j=0}^\infty \exp(-\ell_j^2t)\phi_j(x)\phi_j(y).
\ee
Although $K_t$ satisfies the semigroup property, and
\be\label{heatkernint}
\int_\XX K_t(x,y)d\mu(y) =1, \qquad x\in\XX,
\ee
$K_t$ may  not be the heat kernel in the classical sense. In particular, we do not assume that $K_t$ is nonnegative. The only assumptions we make on $K_t$ are the following: With $\a>0$ as in \eref{ballmeasurecond}, 
\be\label{singlegaussbd}
|K_t(x,y)| \le c_1t^{-\a/2}\exp(-c\rho(x,y)^2/t), \qquad t\in (0,1],\ x,y\in\XX,
\ee
and for any of the first order directional derivatives $\partial$ with respect to a normal coordinate system,
\be\label{heatgradest}
|\partial_y K_t(x,y)| \le c_1t^{-\a/2-1}\exp(-c\rho(x,y)^2/t), \qquad t\in (0,1],\ x,y\in\XX.
\ee
We note that our assumptions imply that $K_t(x,y)$ is well defined for all $x, y\in\XX$ and $t\in (0,1]$. It is proved in \cite{filbirbern} that \eref{singlegaussbd} implies that
\be\label{christupbd}
\sum_{\ell_j\le L}\phi_j^2(x)\le cL^\a, \qquad L>0.
\ee
In the case when $\phi_k$'s (respectively, $\ell_k$'s) are the eigenfunctions (respectively, eigenvalues) of the square root of the negative Laplacian on $\XX$, the assumptions \eref{singlegaussbd} and \eref{heatgradest} can be deduced from the bounds on the spectral functions $\sum_{\ell_j\le L}\phi_j^2(x)$, $\sum_{\ell_j\le L}(\partial\phi_j)^2(x)$ proved by Bin Xu \cite{binxu} (cf. \cite{filbirbern}), and the finite speed of wave propagation. Kordyukov \cite{kordyukov91} has proved similar estimates in the case when $\XX$ has bounded geometry, and $\phi_k$'s are eigenfunctions of a general, second order, strictly elliptic partial differential operator. Other examples, where $\mu$ is not the Riemannian measure on $\XX$ are given by Grigor\'yan in \cite{grigoryanheatmetric}. 

The bounds on the heat kernel are closely connected with the measures of the balls $B(x,r)$. For example, it is  proved in \cite{grigoryanheatmetric}  that the conditions \eref{heatkernint}, \eref{ballmeasurecond}, and  \eref{singlegaussbd} imply   that
\be\label{ballmeasurelowbd}
\mu(B(x,r))\ge cr^\alpha, \qquad 0<r\le 1, \ x\in\XX.
\ee
In view of  \eref{ballmeasurecond}, this shows that $\mu$ satisfies the homogeneity condition
\be\label{doublingcond}
\mu(B(x,R))\le c(R/r)^\a\mu(B(x,r)), \qquad x\in\XX, r\in (0,1],\ R>0.
\ee
In many of the examples cited above, the kernel $K_t$ also satisfies a lower bound to match the upper bound in \eref{singlegaussbd}. In this case, Grigory\'an \cite{grigoryanheatmetric} has also shown that \eref{ballmeasurecond} is satisfied.

 In the case when $\XX$ is the Euclidean sphere, or  the rotation group $SO(3)$, the eigenfunctions of the Laplace--Beltrami operator are polynomials, and hence, if $\Pi_L$ is span of the appropriate eigenfunctions, $P_1,P_2\in\Pi_L$ imply that $P_1P_2\in\Pi_{2L}$. We are not aware of any concrete examples where this is not true. In general,  when $\P_L$ is a span of eigenfunctions of certain elliptic operators, we do not expect such a precise inclusion. Nevertheless,  each of the products $\phi_j\phi_k$ is infinitely often differentiable in this case, and hence, it is reasonable to expect that $\dist(\infty;\phi_j\phi_k,\Pi_m)\to 0$ faster than any polynomial in $1/m$ as $m\to\infty$. Since we are considering an even more  general situation, where $\phi_j$, $\phi_k$ are not assumed to be eigenfunctions of any elliptic operator, we need to make the following assumption as our substitute for the lack of an algebra structure on $\Pi_\infty$. \\
\textsc{Product assumption:}\\
Let $A\ge 2$ be a fixed number, and for $L>0$,
\be\label{gammanormdef}
\e_L:=\sup_{\ell_j,\ell_k\le L}\dist(\infty;\phi_j\phi_k,\Pi_{AL}).
\ee
We assume that $L^c\e_L\to 0$ as $L\to\infty$ for every $c>0$.  We conjecture that if $\XX$ is an analytic manifold and $\phi_j$'s are eigenfunctions of elliptic partial differentiable operators with analytic coefficients, then $\limsup_{L\to\infty}\e_L^{1/L} <1$.

To summarize, our assumptions on the manifold, the measure, and the systems $\{\phi_k\}$, $\{\ell_k\}$
 are: \eref{ballmeasurecond}, \eref{heatkernint}, \eref{singlegaussbd}, \eref{heatgradest}, and the product assumption.

\subsection{Data sets and weights}\label{datasect}
Let $K\subseteq \XX$ be a compact set, $\C\subset K$ be a finite set, . The \emph{mesh norm} $\delta(\C,K)$ of $\C$ relative to $K$ and the minimal separation $q(\C)$ are defined by
\be\label{meshnormdef}
\delta(\C,K) =\sup_{x\in K}\rho(x,\C), \ q(\C)=\min_{x,y\in\C,\ x\not=y}\rho(x,y).
\ee
To keep the notation simple, we will write $\delta(\C):=\delta(\C,\XX)$. Of particular interest in this paper are sets $\C$ satisfying
\be\label{cuniformity}
\delta(\C)\le 2q(\C).
\ee
The proof of the following proposition shows one way to construct such sets from arbitary finite subsets of $\XX$. Consistent with our policy of presenting all proofs in Section~\ref{proofsect}, this proof will be postponed to the end of this paper.

\begin{prop}\label{datasetprop}
{\rm (a)} If $\C\subset \XX$ is a finite set and $\e>0$, there exists $\tilde C\subseteq \C$ such that $\delta(\tilde\C,\C)\le \e \le q(\tilde\C)$. In particular, for the set $\tilde\C$ obtained with $\e=\delta(\C)$, $\delta(\C)\le \delta(\tilde\C)\le 2\delta(\C)\le 2q(\tilde\C)$.\\
{\rm (b)} If $\C_0\subseteq \C_1\subset\XX$ are finite subsets with $\delta(\C_1)\le (1/2)\delta(\C_0)\le q(\C_0)$, then there exists $\C_1^*$, with $\C_0\subseteq\C_1^*\subseteq \C_1$, such that $\delta(\C_1)\le \delta(\C_1^*)\le 2\delta(\C_1)\le 2q(\C_1^*)$.\\
{\rm (c)} Let $\{\C_m\}$ be a sequence of finite subsets of $\XX$, with $\delta(\C_m)\sim 1/m$, and $C_m\subseteq \C_{m+1}$, $m=1,2,\cdots$. Then there exists a sequence of subsets $\{\tilde\C_m \subseteq \C_m\}$, where, for $m=1,2,\cdots,$ $\delta(\tilde\C_m)\sim 1/m$, $\tilde \C_m\subseteq \tilde \C_{m+1}$, $\delta(\tilde\C_m)\le 2q(\tilde \C_m)$. 
\end{prop}

In the sequel, for any finite subset $\C$ (respectively, $\C_m$), we will only work with the subset $\tilde\C$ (respectively, $\tilde\C_m$) as constructed above. Since the rest of the points in $\C$ (respectively, $\C_m$) are ignored in our analysis, we may rename this subset again as $\C$ (respectively, $\C_m$) and assume that $\C$ (respectively, $\C_m$) satisfies \eref{cuniformity}.

The following theorem is proved in \cite{filbirbern}, where do not need the product assumption.
\begin{theorem}\label{mztheo}
  Let $\C$ be a finite subset of $\XX$ (satisfying \eref{cuniformity}), $\delta(\C)\le 1/6$. We assume further that \eref{ballmeasurecond}, \eref{heatkernint}, \eref{singlegaussbd}, and \eref{heatgradest} hold. Then there exists $c>0$ such that for $L \le c\delta(\C)^{-1}$, we have
\be\label{mzineq}
\|P\|_1\le 2\sum_{x\in\C}\mu(B(x,\delta(\C)))|P(x)|\le c_1\|P\|_1, \quad P\in \Pi_L.
\ee
Consequently, for $L \le c\delta(\C)^{-1}$, there exist  numbers $w_x$, $x\in\C$, such that for each $x\in\C$, 
\be\label{wtbds}
|w_x|\le c_2\mu(B(x,\delta(\C)))\le c_3\delta(\C)^\alpha \le c_4q(\C)^\a,
\ee
 and
\be\label{quadrature}
\int_\XX P(y)d\mu(y)=\sum_{x\in\C}w_xP(x), \qquad P\in \Pi_L.
\ee
\end{theorem}

A simple way to find the weights $w_x$ is to solve the least square problem of minimizing $\sum w_x^2$ with the constraints $\sum_{x\in\C}w_x\phi_k(x)=\int_\XX \phi_kd\mu$, $k=0,\cdots,L$ \cite{quadconst}. Alternately, one may obtain $w_x$'s so as to minimize 
$$
\sum_{\ell_k\le L} \left(\sum_{x\in\C}w_x\phi_k(x)-\int_\XX \phi_kd\mu\right)^2.
$$
Efficient numerical algorithms for computing the weights in the context of the unit sphere can be found, for example, in \cite{quadconst, potts, fithe2}. Some of these ideas can be adopted in this context, but our main focus in this paper is of a theoretical nature, and we will not comment further on this issue in this paper.

In view of \eref{ballmeasurelowbd}, \eref{ballmeasurecond}, the inequalities \eref{mzineq} can be formulated as 
\be\label{mzineqdetail}
\|P\|_1\le c_1q(\C)^{\a}\sum_{x\in\C}|P(x)| \le  c_2\delta(\C)^{\a}\sum_{x\in\C}|P(x)|\le c_3\sum_{x\in\C}\mu(B(x,\delta(\C)))|P(x)|\le c_4\|P\|_1, \qquad P\in \Pi_L.
\ee
Inequalities of this nature were proved in the trigonometric case by Marcinkiewicz and Zygmund \cite[Chapter X, Theorem 7.28]{zygmund}. For this reason, we will refer to \eref{mzineqdetail} as MZ inequalities. 

\begin{definition}\label{regularmeasuredef} 
Let $\C\subset\XX$ be a finite set, $a_y$, $y\in\C$ be real numbers, and $d>0$. We will say that $\{a_y\}$ is $d$--regular if for some constant $c$ depending only on $\XX$ and the related quantities described in Section~\ref{manifoldsect}, but not on $\C$, $r$, or $d$, such that
\be\label{mzcond1}
\sum_{y\in\C\cap B(x,r)}|a_y| \le c\{\mu(B(x,r))+d^\a\}, \qquad x\in\XX, \ r>0.
\ee
If $L>0$, we will say that $\{a_y\}$ is a set of quadrature weights (or equivalently, $a_y$'s are quadrature weights) of order $L$ corresponding to $\C$ if
$$
\int_\XX P(y)d\mu(y)=\sum_{y\in\C}a_yP(y), \qquad P\in \Pi_L.
$$
\end{definition}
Thus, for example, the set $\{w_x\}_{x\in\C}$ constructed in Theorem~\ref{mztheo} is a $1/L$--regular set of quadrature weights of order $L$ corresponding to $\C$. We will show in Lemma~\ref{gtildeloclemma} below that the sets $\{a_y\}_{y\in\C}$, where each $a_y=\mu(B(y, \delta(\C)))$ (respectively, $\delta(\C)^\a$, $q(\C)^\a$) are all $\delta(\C)$-- or 
$q(\C)$--regular, but of course, not quadrature weights.

\subsection{Eignets}\label{eignetsect}
 The notion of eignets, analogous to the notion of radial basis function (RBF)/neural networks, is defined  as follows.

\begin{definition}\label{eignetdef} Let $\C\subset \XX$ be a finite set, and $G:\XX\times\XX\to\RR$.  An eignet with centers $\C$ and kernel $G$ is a function of the form $\sum_{y\in\C} a_y G(\circ,y)$, where the coefficients $a_y\in\RR$, $y\in\C$. The set  of all eignets with centers $\C$ will be denoted by $\G(\C)=\G(G;\C)$.
\end{definition}
We note that $\G(\C)$ is a linear space. In the parlace of the theory of RBF/neural networks, the kernel $G$ may be thought of as the activation function.  

As mentioned in the introduction, we are interested in this paper in the case when  the kernel $G$  admits a formal expansion of the form $G(x,y)=\sum_{j=0}^\infty b(\ell_j)\phi_j(x)\phi_j(y)$, where the coefficients $b(\ell_j)$ behave like $\ell_j^{-\beta}$ for some $\beta>0$. (This is the reason for our terminology ``eignet'', to emphasize the formal expansion in terms of what would usually be eigenfunctions of the Laplace--Beltrami operator on a manifold.) The following definition makes this sentiment more precise. In the sequel, $S>\a$ will be a fixed integer.

\begin{definition}\label{eigkerndef}
Let $\beta\in\RR$. A function $b:\RR\to\RR$ will be called a mask of type $\beta$ if $b$ is an even, $S$ times continuously differentiable function such that for $t>0$, $b(t)=(1+t)^{-\beta}F_b(\log t)$ for some $F_b:\RR\to\RR$ such that  $|\derf{F_b}{k}(t)|\le c(b)$, $t\in\RR$, $k=0,1,\cdots,S$,  and $F_b(t)\ge c_1(b)$, $t\in\RR$.     A function $G:\XX\times\XX\to \RR$ will be called  a kernel of type $\beta$ if it admits a formal expansion $G(x,y)=\sum_{j=0}^\infty b(\ell_j)\phi_j(x)\phi_j(y)$ for some mask $b$ of type $\beta>0$. If we wish to specify the connection between $G$ and $b$, we will write $G(b;x,y)$ in place of $G$.
\end{definition}

We observe that $\lim_{t\to-\infty}F_b(t) =b(0)$ is finite. Further, the definition of a mask of type $\beta$ can be relaxed somewhat, for example, the various bounds on $F_b$ and its derivatives may only be assumed for sufficiently large values of $|t|$ rather than for all $t\in\RR$. If this is the case, one can construct a new kernel by adding a suitable diffusion polynomial (of a fixed degree) to $G$, as is customary in the theory of radial basis functions, and obtain a kernel whose mask satisfies the definition given above. This does not add any new feature to our theory. Therefore, we assume the more restrictive definition as given above.

For a $S$ times continuously differentiable function $F$, we define
$$
\||F\||_S:=\sup_{0\le k\le S, x\in\RR}|\derf{F}{k}(x)|.
$$ 
Let $b$ be a mask of type $\beta\in\RR$. In the sequel, if $L>0$, we will write $b_L(t)=b(Lt)$. It is easy to verify by induction that 
$$
\left|t^k\frac{d^k}{dt^k} ((1+t)^\beta b(t))\right|=\left|t^k\frac{d^k}{dt^k}F_b(\log t)\right|\le c(b)c_2, \qquad t> 0, \ k=0,\cdots,S,
$$
and hence,
\be\label{bderupbd}
\left|t^k\frac{d^k}{dt^k} ((1/L+t)^\beta b_L(t))\right|\le c(b)c_2L^{-\beta}, \qquad t> 0, \ k=0,\cdots,S,\ L>0.
\ee
Since $b(t)^{-1}$ is a mask of type $-\beta$, we record that
\be\label{bderlowbd}
\left|t^k\frac{d^k}{dt^k}((1/L+t)^\beta b_L(t))^{-1}\right|\le c(b)c_2L^{\beta}, \qquad t> 0,\ k=0,\cdots,S, \ L>0.
\ee
Finally, if $g:\RR\to\RR$ is any compactly supported, $S$ times continuously differentiable function, such that $g(t)=0$ on some neighborhood of $0$ then \eref{bderupbd}, \eref{bderlowbd} imply 
\be\label{gtimesbineq}
\||gb_L\||_S\le c(b,g)L^{-\beta}, \qquad \||g/b_L\||_S\le c(b,g)L^\beta, \qquad L\ge 1.
\ee

%3
\bhag{Main results}\label{mainsect}
In the remainder of this paper, we fix a number $\beta>0$, a mask $b$ of type $\beta$, and the corresponding kernel $G$. 
Our main goal in this paper is to construct eignets for approximation of functions in $X^p$ and develop an equivalence theroem for approximation by these. In comparison with the approximation theory paradigm described in the introduction, we choose  $X^p$ as the metric space in which the approximation takes place.  We consider a nested sequence $\{\C_m\}$ of finite subsets of $\XX$, each satisfying \eref{cuniformity}, and such that $q(\C_m)\sim \delta(\C_m)\sim 1/m$, $m=1,2,\cdots$. We let $\V_m$ be the space $\G(\C_m)$. Clearly, $\V_m\subset \V_{m+1}$ for $m=1,2,\cdots$. If $\beta>\a/p'$, we will show in Proposition~\ref{networkkernprop} below that each $\V_m\subset X^p$. Our initial choice of smoothness classes is the following. If $f\in L^1+L^\infty$ and $r\ge 0$, we define formally $(\Delta^*)^rf$ by $\ip{(\Delta^*)^rf}{\phi_k}=(1+\ell_k)^r\ip{f}{\phi_k}$, $k=0,1,\cdots$. Let $W^p_r$ be the class of all $f\in X^p$ such that $(\Delta^*)^rf\in X^p$. 
It is proved in \cite{mauropap} (cf. Proposition~\ref{approxlemma} below) that for $f\in W^p_r$ and $L>0$,
$$
\dist(p; f,\Pi_L)\le cL^{-r}\|(\Delta^*)^rf\|_p.
$$
Thus, our goal is to approximate a diffusion polynomial in $\Pi_L$ by eignets, \emph{keeping track of the errors}.
For this purpose, we need another pseudo-differential operator.

\begin{definition}\label{gderdef}
 The operator ${\cal D}={\cal D}_G$ is defined formally by $\ip{{\cal D}f}{\phi_k}=\ip{f}{\phi_k}/b(\ell_k)$, $k=0,1,\cdots$. 
\end{definition}

Clearly, ${\cal D}_G$ is defined on $\Pi_\infty$, and it is easy to verify the fundamental fact that
\be\label{polyaseignet}
P(x)=\int_\XX ({\cal D}_GP)(y)G(x,y)d\mu(y), \qquad P\in\Pi_\infty, \ x\in\XX.
\ee
Our eignets will be discretizations of the integral above. Thus, if $\C\subset \XX$ is a finite set, and ${\bf W}=\{w_y\}_{y\in\C}$ are some real numbers, we define
\be\label{eignetopdef}
\GG(\C;{\bf W}; P, x):=\GG(G;\C;{\bf W}; P, x):=\sum_{y\in\C} w_y ({\cal D}_GP)(y) G(x,y), \qquad P\in\Pi_\infty, \ x\in\XX.
\ee
We note that $\GG$ defines a linear operator on $\Pi_\infty$. 

Our strategy is to approximate a target function $f\in W^p_r$ first by a diffusion polynomial $P\in\Pi_L$ so that $\|f-P\|_p=\O(L^{-r})$. With a careful choice of $\C$ and ${\bf W}$, we will then show that $\|P-\GG(\C;{\bf W};P)\|_p=\O(L^{-r})$. The results are formulated below as our first theorem. We recall the constant $A\ge 2$ described in the ``product assumption'' in Section~\ref{manifoldsect}.

\begin{theorem}\label{directtheofirst}
Let $\C^*\subset\XX$ be a finite set satisfying \eref{cuniformity}, $L\sim q(\C^*)^{-1}$, ${\bf W}^*$ be a $1/L$--regular set of quadrature weights of order $2AL$ corresponding to $\C^*$. Let $1\le p\le\infty$, $\beta>\a/p'$, $0\le r< \beta$. Let  $f\in W^p_r$, and $P\in\Pi_L$ satisfy  $\|f-P\|_p\le cL^{-r}\|(\Delta^*)^rf\|_p$. Then
\be\label{firstdirectest}
\|f-\GG(\C^*;{\bf W}^*; P)\|_p\le c_1L^{-r}\|(\Delta^*)^rf\|_p.
\ee
\end{theorem}

We comment on the construction of the diffusion polynomial $P$ in the above theorem.  In the sequel, we let $h : \RR\to\RR$ be a fixed, infinitely differentiable,  and even function, nonincreasing on $[0,\infty)$, such that $h(t)=1$ if $|t|\le 1/2$ and $h(t)=0$ if $|t|\ge 1$. We will omit the mention of $h$ from the notation, and all constants $c, c_1,\cdots$ may depend upon $h$. We define
\be\label{canonsigmaopdef}
\sigma_L(f,x):=\sigma_L(h;f,x):=\sum_{k=0}^\infty h(\ell_k/L)\ip{f}{\phi_k}\phi_k(x), \qquad L>0, \ x\in\XX, \ f\in L^1+L^\infty.
\ee
It is proved in \cite{mauropap} (cf. Proposition~\ref{approxlemma} below) that $\|f-\sigma_L(f)\|_p \le cL^{-r}\|(\Delta^*)^rf\|_p$, $L>0$. Thus, if $\ip{f}{\phi_k}$ are known (or can be computed) for $\ell_k\le L$, we may take $\sigma_L(f)$ in place of $P$ in Theorem~\ref {directtheofirst}. However, if $f\in X^\infty$ and only the values of $f$ at finitely many sites $\C$ are known, then we may adopt the following procedure instead. First, we consider $L$ (depending upon $\delta(\C)$) such that Theorem~\ref{mztheo} is applicable, and yields a $1/L$--regular set of quadrature weights ${\bf W}=\{w_y\}_{y\in\C}$ of order $2AL$. We then define
\be\label{discsigmaopfirst}
\sigma_L(\C;  {\bf W};f,x):=\sum_{y\in\C}w_yf(y)\left\{\sum_{k=0}^\infty h(\ell_k/L)\phi_k(y)\phi_k(x)\right\}=\sum_{k=0}^\infty h(\ell_k/L)\left\{\sum_{y\in\C}w_yf(y)\phi_k(y)\right\}\phi_k(x),
\ee
which is similar to $\sigma_L(f)$, except that the inner products $\ip{f}{\phi_k}$ are discretized using the quadrature weights. We will prove in Proposition~\ref{approxlemma} below that
\be\label{discapproxfirst}
\|f-\sigma_L(\C;  {\bf W};f)\|_\infty\le cL^{-r}\{\|f\|_\infty +\|(\Delta^*)^rf\|_{\infty}\}, \qquad f\in W^\infty_r, \ L\ge 1.
\ee
Thus, $\sigma_L(\C;  {\bf W};f)$ can also be used in place of $P$ in Theorem~\ref{directtheofirst} in the case when $p=\infty$ to obtain the bound 
\be\label{discfirstdirectest}
\|f-\GG(\C^*;{\bf W}^*; \sigma_L(\C;  {\bf W};f))\|_\infty\le cL^{-r}\{\|f\|_\infty +\|(\Delta^*)^rf\|_{\infty}\}, \qquad f\in W^\infty_r, \ L\ge 1.
\ee
 We may  choose $\C^*=\C$ and ${\bf W}^*={\bf W}$ in this case, but do not have to do so. On the other hand, if one does not discretize the inner products $\ip{f}{\phi_k}$ so carefully, then the approximation error might be substantially worse than that in \eref{discapproxfirst}, as shown in the case of the sphere in \cite{quadconst}. The eignets $\GG(\C^*;{\bf W}^*;P)$ with these choices of $P$ have the advantage of stability as described in Theorem~\ref{jacksontheofirst} below.

Next, we wish to consider the question whether the estimate \eref{firstdirectest} is the best possible \emph{for individual functions}, and whether the method of approximation described is the best possible. We wish we could say that if there is any sequence $s_m\in \V_m$ of eignets with $\|f-s_m\|_p=\O(m^{-r})$ then necessarily $f\in W^p_r$. However, such a statement is not true even in the classical trigonometric case. For example, for any $r>0$, the function $\disp f(x)=\sum_{k=1}^\infty \frac{\sin kx}{k^{1+r}}$ satisfies the condition that the uniform degree of approximation to $f$ from trigonometric polynomials of degree at most $m$ is $\O(m^{-r})$. However, there is a continuous function $f_1$ such that $\disp (\Delta^*)^r f(x) =f_1(x) +\sum_{k=1}^\infty \frac{\sin kx}{k}$ is not continuous. In the classical trigonometric case, one needs to enlarge the smoothness class to achieve such an equivalence. This is done via $K$-functionals. We now introduce this notion in the present context. Not to confuse the notation with the heat kernel or the corresponding operator, we will use the notation $\omega$ for the $K$-functional, motivated by the equivalence of the $K$--functional and a modulus of smoothness in the trigonometric case.

If $f\in X^p$, $r>0$ is an integer, we define for $\delta>0$
\be\label{kfuncdef}
\omega_r(p;f,\delta):=\inf\{\|f-f_1\|_p +\delta^r\|(\Delta^*)^rf_1\|_{p}\ :\ f_1\in W^p_r\}.
\ee
If $\gamma>0$, we choose an integer $r>\gamma$, and define the smoothness class $H^p_\gamma$ to be the class of all $f\in X^p$ such that 
\be\label{hpgammadef}
\|f\|_{H^p_\gamma}:=\sup_{\delta\in (0,1]}\frac{\omega_r(p;f,\delta)}{\delta^\gamma} <\infty.
\ee
It can be shown that different values of $r>\gamma$ give rise to the same smoothness class with equivalent norms (cf. \cite{devlorbk}). We note that $W^p_r\subset H^p_r$ for every integer $r\ge 1$. The class $H^p_r$ turns out to be the right enlargement for characterization by approximation by eignets. 

First, however, we wish to state the following version of Theorem~\ref{directtheofirst} in the case when the special polynomials are chosen in place of $P$ in that theorem. A popular technique in learning theory is to obtain an approximation by minimizing a regularization functional. For example, the quantity $\omega_r(p;f,\delta)$ is such a functional. The following theorem shows that the operators $\GG$ defined with these special polynomials satisfy, up to a constant multiple, a minimal regularization property.

\begin{theorem}\label{jacksontheofirst}
Let $1\le p\le \infty$, $f\in X^p$, $\beta>\a/p'$, $0<r<\beta-\a/p'$, $L>0$, $\C^*$, ${\bf W}^*$ be as in Theorem~\ref {directtheofirst}. \\
{\rm (a)} With $\GG_L(f,x)=\sigma(\C^*; {\bf W}^*; \sigma_L(f),x)$, $x\in\XX$, we have
\be\label{jacksonestfirst}
\|f-\GG_L(f)\|_p +L^{-r}\|(\Delta^*)^r\GG_L(f)\|_p \le c\omega_r(p;f,1/L).
\ee
In particular, $\|\GG_L(f)\|_p\le c\|f\|_p$. \\
{\rm (b)}Let $\C\subset\XX$ be a finite set satisfying \eref{cuniformity}, ${\bf W}=\{w_y\}_{y\in\C}$ be a $1/L$-regular set of quadrature weights on $\C$ of order $2AL$. For $\tilde\GG_L(\C;{\bf W};f,x)=\sigma(\C^*; {\bf W}^*; \sigma_L(\C;  {\bf W};f),x)$, $x\in\XX$, we have
\be\label{discstability}
\|\tilde\GG_L(\C;{\bf W};f)\|_p\le c\left\{\sum_{y\in\C}|w_y||f(y)|^p\right\}^{1/p},
\ee
and
\be\label{discjacksonestfirst}
\|f-\tilde\GG_L(\C;{\bf W};f)\|_\infty +L^{-r}\|(\Delta^*)^r\tilde\GG_L(\C;{\bf W};f)\|_\infty \le c\{\omega_r(\infty;f,1/L) +L^{-r}\|f\|_\infty\}.
\ee
\end{theorem}

We are now ready to state the equivalence theorem for the spaces $\V_m$ described at the beginning of this section. We assume  that for each $m\ge 1$, $q(\C_m)\sim 1/m$, and there exists a set of $1/m$--regular set ${\bf W}_m$ of quadrature weights of order $2Am$ based on the set $\C_m$. For $1\le p\le\infty$ and $f\in X^p$, let
\be\label{eignetseqdef}
\GG_m(f,x):=\GG(\C_m; {\bf W}_m; \sigma_m(f),x), \qquad x\in\XX, \ m=1,2,\cdots.
\ee
We note that there is no conflict with the notation in Theorem~\ref{jacksontheofirst}, since we may choose $\C^*=\C_L$, ${\bf W}^*={\bf W}_L$.

\begin{theorem}\label{equivtheo}
Suppose that 
\be\label{heatlowbd}
K_t(x,x)\ge ct^{-\a/2}, \qquad x\in\XX,\ t\in (0,1].
\ee
Then the following are equivalent for each $\gamma$ with $0<\gamma<\beta-\a/p'$:\\
{\rm (a)} $f\in H^p_\gamma$.\\
{\rm (b)} $\sup_{m\ge 1}m^{\gamma}\|f-\GG_m(f)\|_p \le c(f)$.\\
{\rm (c)} $\sup_{m\ge 1}m^{\gamma}\dist(L^p;f,\G(\C_m)) \le c(f)$.\\
In the case when $p=\infty$, each of these assertions is also equivalent to \\
{\rm (d)} $\sup_{m\ge 1}m^{\gamma}\|f-\GG(\C_m; {\bf W}_m; \sigma_m(\C_m;{\bf W}_m;f))\|_\infty \le c(f)$.
\end{theorem}
Thus, if one considers the class $H^p_\gamma$ in place of $W^p_r$, then the estimates of the form given in Theorem~\ref{equivtheo} (b) (or (d)) are best possible for individual functions.  One may also formulate a similar equivalence theorem for Besov spaces, defined by replacing the supremum expression in \eref{hpgammadef} by a suitable integral expression. However, this would only complicate our notations rather than adding any new insight into the subject. Therefore, we prefer not to do so. We note that  in the case when $\phi_j$'s (respectively $\ell_j$'s) are the eigenfunctions (respectively, eigenvalues) of the negative square root of the Laplace--Beltrami operator, then  Minakshisundaram  and  Pleijel have proved an asymptotic expression for the heat kernel in \cite{mincanad}, which implies both \eref{heatlowbd} and \eref{singlegaussbd}. In \cite{hormander}, H\"ormander has obtained uniform asymptotics for the sums $\sum_{\ell_j\le L}\phi_j^2(x)$ for a very general class of elliptic differential operators on a manifold. It will be shown in Lemma~\ref{tauberlemma} that these lead to \eref{heatlowbd} and \eref{singlegaussbd} (with $x=y$). Further examples are given by Grigor\'yan \cite{grigoryan99} and references therein.

We end this section by recording two interesting facts, valid for arbitrary eignets of type $\beta$. The first of these facts relates the coefficients of the eignet with its norm. For a sequence (or vector) of complex numbers ${\bf a}=\{a_j\}$ and $1\le p\le\infty$, we denote by $\|{\bf a}\|_{\ell^p}$, the usual sequential (or Euclidean) $\ell^p$ norm. 

\begin{theorem}\label{coefftheo}
We assume that \eref{heatlowbd} holds. Let $1\le p\le\infty$,  $\beta>\a/p'$,  $\C\subset\XX$ be a finite set,  $a_y\in\RR$, $y\in\C$, and ${\bf a}=(a_y)_{y\in\C}$. Then
\be\label{coeffineq}
\|{\bf a}\|_{\ell^p}\le cq(\C)^{\a/p'-\beta}\left\|\sum_{y\in\C} a_yG(\circ,y)\right\|_p.
\ee
\end{theorem}

The second fact describes the simultaneous approximation property of eignets.

\begin{theorem}\label{simapproxtheo}
We assume that \eref{heatlowbd} holds. Let $1\le p\le\infty$,   $0<\gamma< \beta-\a/p'$, $0< \gamma\le r< \beta$, and $f\in W^p_r$. If $\Psi_m\in\V_m$ satisfy $\|f-\Psi_m\|_p\le cm^{-r}\|(\Delta^*)^r f\|_{p}$ then also $\|(\Delta^*)^\gamma f-(\Delta^*)^\gamma \Psi_m\|_p \le cm^{\gamma-r}\|(\Delta^*)^r f\|_{p}$. 
\end{theorem}

%4
\bhag{An abstraction}\label{abstractsect}

In our proofs, we need to estimate many sums and integrals. Since these estimates involve similar ideas, we prefer to deal with them in a unifed manner by treating sums as integrals with respect to finitely supported measures.  We observe that if $\C\subset\XX$, and $W_x$, $x\in\C$, are any real numbers,
 a sum of the form $\sum_{x\in \C} W_x f(x)$ can be
expressed as a Lebesgue--Stieltjes integral $\int fd\nu$, where $\nu$ is the measure that
associates the mass $W_x$ with each point $x\in\C$. The total variation measure in
this case is given by $|\nu|(B)=\sum_{x\in B\cap \C}|W_x|$, $B\subset \XX$. Thus, for example, in \eref{discsigmaopfirst}, if $\nu$ is the measure that associates with each $y\in \C$ the mass $w_y\in {\bf W}$, then we may write
\be\label{discsigmaopdef}
\sigma_L(\nu;f,x):=\int_\XX f(y)\sum_{k=0}^\infty h(\ell_k/L)\phi_k(y)\phi_k(x)d\nu(y)
\ee
in place of the more cumbersome notation $\sigma_L(\C;{\bf W}; f,x)$, helping us thereby to focus our attention on the essential aspects of this measure rather than the choice of $\C$ and $\W$. Moreover, if one takes $\mu$ in place of $\nu$, then $\sigma_L(\mu;f)=\sigma_L(f)$. 
In addition to being concise, this notation has another major advantage. If the information available about the target function $f$ is neither the spectral data $\{\ip{f}{\phi_k}\}$ nor point evaluations, but, for example, averages of $f$ over small balls, the notation allows one to treat this case as well without introducing yet another notation, just by defining $\nu$ appropriately. In the sequel, with the exception of a few occasions, we will typically use $\nu$ to be one of the following measures: (1) $\mu$, (2) the measure that associates the mass $w_y$ with each $y\in\C$ for some $\C$,  (3) the measure that associates the mass $q(\C)^\alpha$ with each $y\in\C$,  and (4) various linear combinations of the above measures. 

To demonstrate a technical advantage, Definition~\ref{regularmeasuredef} takes the following form, where the ambiguity and tacit understanding about what the constants depend upon can be avoided, and we get the full advantage of the vector space properties of measures.
\begin{definition}\label{absregularmeasuredef}
Let $d>0$. A signed measure $\nu$ defined on $\XX$ will be called  $d$--regular if there exists a constant $c=c(\nu)>0$ such that
\be\label{mzmeasdef}
 |\nu|(B(x,r))\le c\left\{\mu(B(x,r))+d^{\a}\right\}, \qquad x\in\XX, \ r>0,
\ee
where $\a$ is the constant introduced in \eref{ballmeasurecond}.
Let ${\cal M}_d$ denote the class of all signed measures satisfying \eref{mzmeasdef}. Then ${\cal M}_d$ is a vector space. For $\nu\in {\cal M}_d$, if we denote by $\|\nu\|_{{\cal M}_d}$ the infimum of $c$ which serves in \eref{mzmeasdef}, then $\|\circ\|_{{\cal M}_d}$ is a norm on ${\cal M}_d$.
\end{definition}
For example, $\mu$ itself is in  ${\cal M}_d$ with $\|\mu\|_{{\cal M}_d}=1$ for \emph{every} $d>0$. If $\C\subset\XX$ is as in Theorem~\ref{mztheo}, then we will show in Lemma~\ref{gtildeloclemma} below that the measures that associate the mass $\mu(B(x,\delta(\C)))$ (respectively, $\delta(\C)^\a$, $q(\C)^\a$, $w_x$, $|w_x|$) with $x\in \C$ are all in ${\cal M}_{\delta(\C)}$ as well as ${\cal M}_{q(\C)}$ with $\|\nu\|_{{\cal M}_{q(\C)}}\le c$, where the constant is independent of $\C$. It is also easy to see that for any $c>0$, ${\cal M}_d\subseteq {\cal M}_{cd}$, with $\|\nu\|_{{\cal M}_{cd}}\le \max(1,c^\a)\|\nu\|_{{\cal M}_d}$.
In view of \eref{ballmeasurecond} and \eref{ballmeasurelowbd}, the condition \eref{mzmeasdef} is equivalent to
\be\label{mzmeasequdef}
|\nu|(B(x,r))\le c\|\nu\|_{{\cal M}_d} (r+d)^\a\le c_1\|\nu\|_{{\cal M}_d}\mu(B(x,r+d)).
\ee
Finally, we note that since $\mu$ is a probability measure, the condition \eref{mzmeasdef} implies that $|\nu|(B)\le c(1+d^\a)$ for every ball $B\subset \XX$, and hence, that $|\nu|(\XX)\le c(1+d^\a)$ as well.

The quadrature formula \eref{quadrature} can be restated in the form
\be\label{quadmeasure}
\int_\XX P(y)d\mu(y)=\int_\XX P(y)d\nu(y), \qquad P\in \Pi_L,
\ee
where $\nu$ is the measure that associates the mass $w_y$ with each $y\in\C$. Any  (signed or positive) measure $\nu$ satisfying \eref{quadmeasure} will be called a quadrature measure of order $L$; in particular, $\mu$ itself is a quadrature measure of order $L$ for every $L>0$.

If $\nu$ is a signed or positive Borel measure on $\XX$, $X\subseteq \XX$ is $\nu$-measurable, and $f : X\to \CC$ is a $\nu$-measurable function, we will write
$$
\|f\|_{\nu;X,p}:=\left\{\begin{array}{ll}
\disp \left\{\int_{X}|f(x)|^pd|\nu|(x)\right\}^{1/p}, & \mbox{ if $1\le p<\infty$,}\\
|\nu|-\esssup_{x\in X}|f(x)|, &\mbox{ if $p=\infty$.}
\end{array}\right.
$$
We will write $L^p(\nu;X)$ to denote the class of all $\nu$--measurable functions $f$ for which $\|f\|_{\nu;X,p}<\infty$, where two functions are considered equal if they are equal $|\nu|$--almost everywhere. To make the notation consistent with the one introduced before, we will omit the mention of $\nu$ if $\nu=\mu$ and that of $X$ if $X=\XX$. 

 In the sequel, for any $H:\RR\to\RR$, we define formally
\be\label{phikerndef}
\Phi_L(H;x,y):=\sum_{j=0}^\infty H(\ell_j/L)\phi_j(x)\phi_j(y), \qquad x,y\in\XX, \ L>0.
\ee
For example, $G(x,y)=\Phi_L(b_L;x,y)$.
If $\nu$ is any measure on $\XX$ and $f\in L^p$, we may define formally
\be\label{sigmaopdef}
\sigma_L(H;\nu; f,x):=\int_\XX f(y)\Phi_L(H;x,y)d\nu(y). 
\ee
As before, we will omit the mention of $\nu$ if $\nu=\mu$ and that of $H$ if  $H=h$. Thus, $\Phi_L(x,y)=\Phi_L(h;x,y)$, and similarly  $\sigma_L(f,x)=\sigma_L(h;\mu;f)$, $\sigma_L(\nu;f,x)=\sigma_L(h;\nu;f,x)$. The slight inconsistency is resolved by the fact that we use $\mu$, $\nu$, $\tilde\nu$ etc. to denote measures,  $h$, $g$, $b$, $H$, etc. to denote functions, and $X$, $\XX$ to denote sets. We do not consider this to be a sufficiently important issue to complicate our notations. We note that $\sigma_L(G(\circ,y),x)=\Phi_L(hb_L;x,y)$. 

In the sequel, we define $g$ by $g(t)=h(t)-h(2t)$. We note that $g$ is supported on $(1/4,1)\cup (-1,-1/4)$, and
\be\label{gsum}
h\left(\frac{t}{2^n}\right)= h(t)+\sum_{k=1}^n g\left(\frac{t}{2^k}\right), \qquad t\in\RR,\ n=1,2,\cdots.
\ee

%5
\bhag{Technical preparation}\label{techsect}
 In Section~\ref{kernsect}, we prove a few facts regarding the kernels $\Phi_L$, which will be used very often in the proofs in Section~\ref{proofsect} as well as the rest of the proofs in this section. In Section~\ref{diffpolysect}, we describe several properties of diffusion polynomials and approximation by these. Since we do not need all the assumptions listed in Section~\ref{manifoldsect}, we will list in each theorem only those assumptions which are needed there.

\subsection{Kernels}\label{kernsect}

We will often use the following simple application of the Riesz--Thorin interpolation theorem \cite[Theorem~1.1.1]{bergh} to estimate the operators defined in terms of kernels. 

\begin{lemma}\label{rieszthorinlemma}
Let $\nu_1$, $\nu_2$ be signed measures (having bounded variation) on a measure space $\Omega$, supported on  $\Omega_1$ and $\Omega_2$ respectively, $\Phi : \Omega\times\Omega\to\RR$ be a bounded, $|\nu_1|\times|\nu_2|$ measurable function,  $1\le p\le\infty$, $f\in L^p(|\nu_1|)$, and let
$$
T_f(x):=\int f(t)\Phi(x,t)d\nu_1(t).
$$
Then with
$$
A_1 = \sup_{t\in \Omega_1}\|\Phi(\cdot,t)\|_{|\nu_2|;\Omega,1},\ A_\infty=\sup_{x\in \Omega_2}\|\Phi(x,\cdot)\|_{|\nu_1|;\Omega,1},
$$
we have
\be\label{rieszthorinest}
\|T_f\|_{|\nu_2|;\Omega, p}\le A_1^{1/p}A_\infty^{1/p'}\|f\|_{|\nu_1|;\Omega, p}.
\ee
\end{lemma}
\begin{Proof} It is clear that $\|T_f\|_{|\nu_2|;\Omega,\infty}\le A_\infty\|f\|_{|\nu_1|;\Omega, \infty}$. Fubini's theorem can be used to see that $\|T_f\|_{|\nu_2|;\Omega,1}\le A_1\|f\|_{|\nu_1|;\Omega, 1}$. The estimate \eref{rieszthorinest} follows by Riesz--Thorin interpolation theorem. 
\end{Proof}

The starting point of our proofs is to recall the following theorem proved in \cite{mauropap}, and in \cite{filbirbern} in somewhat greater generality, stating the assumptions as they are stated in this paper.

\begin{theorem}\label{kerntheo}
Let  $S>\a$ be an integer, $H:\RR\to \RR$ be an even, $S$ times continuously differentiable function, supported on $[-1,1]$. 
We assume further that \eref{ballmeasurecond}, \eref{singlegaussbd} hold. Then for every $x,y\in \XX$, $L>0$,
\be\label{kernlocest}
| \Phi_L(H;x,y)|\le \frac{cL^{\a}\||H\||_S}{\max(1, (L\rho(x,y))^S)}.
\ee
Consequently,
\be\label{kernbdest}
\sup_{x\in\XX}\int_{\XX}| \Phi_L(H;x,y)|d\mu(y) \le c\||H\||_S,
\ee
and for every $1\le p\le \infty$ and $f\in L^p$, 
\be\label{sigmaopbd}
\| \sigma_L(H;f)\|_p\le c\||H\||_S\|f\|_p.
\ee
\end{theorem}

The following Propositions~\ref{criticalprop} and \ref{networkkernprop} will be used very often in this section, with different interpretations for $H$ and the measures involved.

\begin{prop}\label{criticalprop}
Let $d>0$, $S$, $H$ be as in Theorem~\ref{kerntheo}, and \eref{ballmeasurecond}, \eref{singlegaussbd} hold. Let $\nu\in {\cal M}_d$,  $L>0$,  and $c$ be the constant that appears in \eref{ballmeasurecond}. Let $1\le p\le \infty$, $1/p'+1/p=1$.\\
{\rm (a)} If $g_1:[0,\infty)\to [0,\infty)$ is a nonincreasing function, then for any $L>0$, $r>0$, $x\in\XX$,
\be\label{g1ineq}
L^\a\int_{\Delta(x,r)}g_1(L\rho(x,y))d|\nu|(y)\le \frac{2^{\a}(c+(d/r)^\a)\a}{1-2^{-\a}}\|\nu\|_{{\cal M}_d}\int_{rL/2}^\infty g_1(u)u^{\a-1}du.
\ee
{\rm (b)} If  $r\ge  1/L$, then
\be\label{phiintaway}
\int_{\Delta(x,r)}|\Phi_L(H;x,y)|d|\nu|(y) \le c_1(1+(dL)^\a)(rL)^{-S+\a}\|\nu\|_{{\cal M}_d}\||H\||_S.
\ee
{\rm (c)} We have
\be\label{phiinttotal}
\int_\XX|\Phi_L(H;x,y)|d|\nu|(y)\le c_2\{(1+(dL)^\a)\}\|\nu\|_{{\cal M}_d}\||H\||_S,
\ee 
\be\label{philpnorm}
\|\Phi_L(H;x,\circ)\|_{\nu;\XX,p} \le c_3L^{\a/p'}\{(1+(dL)^\a)\}^{1/p}\|\nu\|_{{\cal M}_d}\||H\||_S.
\ee
\end{prop}
\begin{Proof}\ %of Proposition~\ref{criticalprop}
By replacing $\nu$ by $|\nu|/\|\nu\|_{{\cal M}_d}$, we may assume that $\nu$ is  positive, and $\|\nu\|_{{\cal M}_d}=1$. With a similar normalization with $H$, we may also assume that $\||H\||_S=1$. Moreover, for  $r>0$, $\nu(B(x,r))\le \mu(B(x,r))+d^\a \le (c+(d/r)^\a)r^\a$, where $c$ is the constant appearing in \eref{ballmeasurecond}. In this proof only, we will write ${\cal A}(x,t)=\{y\in\XX\ :\ t< \rho(x,y)\le 2t\}$. We note that $\nu({\cal A}(x,t))\le 2^\a(c+(d/r)^\a)t^\a$, $t\ge r$, and 
$$
\int_{2^{R-1}}^{2^R}u^{\a-1}du = \frac{1-2^{-\a}}{\a} 2^{R\a}.
$$
Since $g_1$ is nonincreasing, we have
\begin{eqnarray*}
\lefteqn{\int_{\Delta(x,r)}g_1(L\rho(x,y))d\nu(y)=\sum_{R=0}^\infty \int_{{\cal A}(x,2^{R}r)}g_1(L\rho(x,y))d\nu(y)}\\
&\le& \sum_{R=0}^\infty g_1(2^RrL)\nu({\cal A}(x,2^{R}r)) \le 2^\a(c+(d/r)^\a)\sum_{R=0}^\infty g_1(2^RrL)(2^{R}r)^\a\\
&\le&\frac{2^{\a}(c+(d/r)^\a)\a}{1-2^{-\a}}r^\a\sum_{R=0}^\infty \int_{2^{R-1}}^{2^R} g_1(urL)u^{\a-1}du =\frac{2^{\a}(c+(d/r)^\a)\a}{1-2^{-\a}}r^\a\int_{1/2}^\infty g_1(urL)u^{\a-1}du\\
&=&\frac{2^{\a}(c+(d/r)^\a)\a}{1-2^{-\a}}L^{-\a}\int_{rL/2}^\infty g_1(v)v^{\a-1}dv.
\end{eqnarray*}
This proves \eref{g1ineq}.

Let $x\in\XX$, $L>0$. For $r\ge 1/L$, $d/r\le dL$. In view of \eref{kernlocest} and \eref{g1ineq}, we have for $x\in \XX$:
\begin{eqnarray*}
\int_{\Delta(x,r)}|\Phi_L(H;x,y)|d\nu(y)&\le& c_1L^\a \int_{\Delta(x,r)}(L\rho(x,y))^{-S}d\nu(y)\le 
c_1(c+(dL)^\a)\int_{rL/2}^\infty v^{-S+\a-1}dv\\
&\le&c_2(1+(dL)^\a)(rL)^{-S+\a}.
\end{eqnarray*}
This proves \eref{phiintaway}.

Using \eref{phiintaway} with $r=1/L$, we obtain that
\be\label{pf1eqn1}
\int_{\Delta(x,1/L)}|\Phi_L(H;x,y)|d\nu(y)\le c_2(1+(dL)^\a).
\ee
We observe that in view of \eref{kernlocest}, and the fact that $\nu(B(x,1/L))\le c_1(1/L+d)^\a \le c_1L^{-\a}(1+(dL)^\a)$,
$$
\int_{B(x,1/L)}|\Phi_L(H;x,y)|d\nu(y)\le c_1L^\a\nu(B(x,1/L))\le c_1(1+(dL)^\a).
$$
 Together with \eref{pf1eqn1}, this leads to \eref{phiinttotal}.

The estimate \eref{philpnorm} follows from \eref{kernlocest} in the case $p=\infty$, and from \eref{phiinttotal} in the case $p=1$. For $1<p<\infty$, it follows from the convexity inequality
\be\label{convexityineq}
\|F\|_{\nu;\XX,p}\le \|F\|_{\nu;\XX,\infty}^{1/p'}\|F\|_{\nu;\XX,1}^{1/p}.
\ee
\end{Proof}

\begin{cor}\label{btimeshcor}
Let $\beta\in\RR$, $\tilde b$ be a mask of type $\beta$,   $n\ge 1$ be an integer,  $\nu\in{\cal M}_{2^{-n}}$, and \eref{ballmeasurecond}, \eref{singlegaussbd} hold. Then for integer $n\ge 1$,
\be\label{btimeshint}
\sup_{x\in\XX}\int_\XX|\Phi_{2^n}(h\tilde b_{2^n};x,y)|d|\nu|(y)\le c\|\nu\|_{{\cal M}_{2^{-n}}}\left\{\begin{array}{ll}
2^{-n\beta}, &\mbox{ if $\beta<0$,}\\
n, &\mbox{ if $\beta=0$,}\\
1, &\mbox{ if $\beta>0$,}
\end{array}\right.
\ee
and for $1\le p\le\infty$,
\be\label{btimeshnorm}
\|\Phi_{2^n}(h\tilde b_{2^n};x,\circ)\|_p \le c\|\nu\|_{{\cal M}_{2^{-n}}}\left\{\begin{array}{ll}
2^{-n(\beta-\a/p')}, &\mbox{ if $\beta<\a/p'$,}\\
n, &\mbox{ if $\beta=\a/p'$,}\\
1, &\mbox{ if $\beta>\a/p'$.}
\end{array}\right.
\ee
\end{cor}
\begin{Proof}\ %of Corollary~\ref{btimeshcor}
We normalize $\nu$ so that $\|\nu\|_{{\cal M}_{2^{-n}}}=1$.
In view of \eref{gsum},
\bea\label{pf12eqn1}
\lefteqn{\Phi_{2^n}(h\tilde b_{2^n};x,y)=\sum_{j=0}^\infty h\left(\frac{\ell_j}{2^n}\right)\tilde b(\ell_j)\phi_j(x)\phi_j(y)}\nonumber\\
&=& \sum_{\ell_j\le 1} h(\ell_j)\tilde b(\ell_j)\phi_j(x)\phi_j(y)+\sum_{k=1}^n \sum_{j=0}^\infty g\left(\frac{\ell_j}{2^k}\right)\tilde b(\ell_j)\phi_j(x)\phi_j(y)\nonumber\\
&=& \sum_{\ell_j\le 1} h(\ell_j)\tilde b(\ell_j)\phi_j(x)\phi_j(y)+\sum_{k=1}^n \Phi_{2^k}(g\tilde b_{2^k};x,y).
\eea
Since $h$ and $\tilde b$ are both bounded functions,  \eref{christupbd} shows that
\be\label{pf12eqn2}
\left|\sum_{\ell_j\le 1} h(\ell_j)\tilde b(\ell_j)\phi_j(x)\phi_j(y)\right|\le c, \qquad x,y\in\XX.
\ee
In view of \eref{gtimesbineq} used with $\tilde b$ in place of $b$, and \eref{phiinttotal} used with $d=2^{-n}$, $L=2^k$, $H=g\tilde b_{2^k}$, we obtain
$$
\sup_{x\in\XX}\int_\XX |\Phi_{2^k}(g\tilde b_{2^k};x,y)|d|\nu|(y)\le c2^{-k\beta}, \qquad k=1,2,\cdots,n.
$$
Together with \eref{pf12eqn1} and \eref{pf12eqn2}, this leads to \eref{btimeshint}. The proof of \eref{btimeshnorm} is similar; we use \eref{philpnorm} in place of \eref{phiinttotal}.
\end{Proof}

We observe that if $\C$ is a finite subset of $\XX$,  $\nu$ is the measure that associates the mass $q(\C)^\a$ with each $y\in\C$, then an eignet $\Psi(x) =\sum_{y\in \C} a_yG(x,y)$ can be expressed as $q(\C)^{-\a}\int_\XX a(y)G(x,y)d\nu(y)$, and $\sigma_L(\Psi,x)=q(\C)^{-\a}\int_\XX a(y)\Phi_L(hb_L;x,y)d\nu(y)$. One of the applications of the following proposition is then to estimate $\|\Psi-\sigma_L(\Psi)\|_p$. A different application is given in Lemma~\ref{polytoeignetlemma}.

\begin{prop}\label{networkkernprop}
Let $1\le p\le\infty$, $\beta>\a/p'$,  $b$ be a mask of type $\beta$, and \eref{ballmeasurecond}, \eref{singlegaussbd} hold.\\
{\rm (a)} For every $y\in\XX$, there exists $\psi_y:=G(\circ,y)\in X^p$ such that $\ip{\psi_y}{\phi_k} =b(\ell_k)\phi_k(y)$, $k=0,1,\cdots$. We have
\be\label{glpuniform}
\sup_{y\in\XX}\|G(\circ,y)\|_p \le c.
\ee
{\rm (b)} Let 
 $n\ge 1$ be an integer, $\nu\in{\cal M}_{2^{-n}}$, and for $F\in L^1(\nu)\cap L^\infty(\nu)$, $m\ge n$,
$$
U_m(F,x):=\int_{y\in\XX} \{G(x,y)-\Phi_{2^{m}}(hb_{2^{m}};x,y)\}F(y)d\nu(y).
$$
Then 
\be\label{terrest}
\|U_m(F)\|_p \le c2^{-m\beta}2^{\a(m-n)/p'}\|\nu\|_{{\cal M}_{2^{-n}}}\|F\|_{\nu;\XX,p}.
\ee
\end{prop}

\begin{Proof}\ % of Proposition~\ref{networkkernprop}.}

Since $\mu\in {\cal M}_d$  and $\|\mu\|_{{\cal M}_d}=1$ for every $d>0$, we conclude from \eref{gtimesbineq} and \eref{philpnorm} (used with $\mu$ in place of $\nu$, $1/L$ in place of $d$, $H=gb_{L}$),  that
$$
\sup_{y\in\XX}\|\Phi_{L}(gb_{L};y,\circ)\|_p \le cL^{\a/p'-\beta}, \qquad L\ge 1.
$$
Since $\beta>\a/p'$, we conclude  for integers $1\le n\le N$,
\be\label{pf7eqn1}
\sup_{y\in\XX}\left\|\sum_{j=n+1}^N \Phi_{2^{j}}(gb_{2^{j}};y,\circ)\right\|_p
\le \sum_{j=n+1}^N \sup_{y\in\XX}\|\Phi_{2^{j}}(gb_{2^{j}};y,\circ)\|_p \le c2^{n(\a/p'-\beta)}.
\ee
Thus, the sequence
\be\label{pf7eqn2}
\Phi_1(hb_1;y,\circ) +\sum_{j=1}^n \Phi_{2^{j}}(gb_{2^{j}};y,\circ) =\Phi_{2^n}(hb_{2^n}; y,\circ)
\ee
 converges in $L^p$ to some function in $X^p$, uniformly in $y$. Denoting this function by $\psi_y$, it is easy to calculate that $\ip{\psi_y}{\phi_k} =b(\ell_k)\phi_k(y)$. Thus, the formal expansion of $\psi_y$ is the same as that of $G(\circ,y)$. Moreover,
$$
\sigma_{2^n}(\psi_y,x)=\sigma_{2^n}(G(x,\circ),y) = \Phi_{2^n}(hb_{2^n}; y,x) 
$$
converges to $G(x,y)$ in the sense of $L^p$ in $x$, and uniformly in $y$. The estimate \eref{glpuniform} is clear from \eref{pf7eqn1} and \eref{pf7eqn2}.

To prove part (b), we use a similar argument again.
 Without loss of generality, we may assume that $\nu$ is a positive measure and $\|\nu\|_{{\cal M}_{2^{-n}}}=1$. 
Let $j\ge n$ be an integer. Using \eref{gtimesbineq}, \eref{phiinttotal} with $2^{-n}$ for $d$, $2^j$ in place of $L$, and oberving that $dL\ge 1$ with these choices, we obtain
\be\label{pf3eqn1}
\sup_{x\in\XX}\int_\XX|\Phi_{2^{j}}(gb_{2^{j}};x,y)|d\nu(y) \le c2^{-n\a}2^{-j(\beta-\a)}.
\ee
Using \eref{gtimesbineq}, \eref{phiinttotal} with $\mu$ in place of $\nu$,  $2^j$ in place of $L$, and $2^{-j}$ for $d$, we obtain
$$
\sup_{x\in\XX}\int_\XX|\Phi_{2^{j}}(gb_{2^{j}};x,y)|d\mu(y) \le c2^{-j\beta}.
$$
Hence, Lemma~\ref{rieszthorinlemma} with $\nu$ in place of $\nu_1$, $\mu$ in place of $\nu_2$, implies that
\be\label{pf3eqn3}
\left\|\int_\XX \Phi_{2^{j}}(gb_{2^{j}};\circ,y)F(y)d\nu(y)\right\|_p\le c2^{-n\a/p'}2^{-j(\beta-\a/p')}\|F\|_{\nu;\XX,p}.
\ee
Since $\beta>\a/p'$, the sequence
$$
\int_\XX \Phi_{2^n}(hb_{2^n};\circ,y)F(y)d\nu(y) =\int_\XX \Phi_1(hb;\circ,y)F(y)d\nu(y)+\sum_{j=1}^n \int_\XX \Phi_{2^{j}}(gb_{2^{j}};\circ,y)F(y)d\nu(y)
$$
converges in the sense of $L^p$ to some function in $X^p$. Since $\Phi_{2^n}(hb_{2^n};\circ,y)\to G(\circ,y)$ in the sense of $L^p$  uniformly in $y$, this function must be $\int_\XX G(\circ,y)F(y)d\nu(y)$. Consequently,
$$
U_m(F,\circ) = \sum_{j=m+1}^\infty \int_\XX \Phi_{2^{j}}(gb_{2^{j}};\circ,y)F(y)d\nu(y)
$$
in the sense of $L^p$, and \eref{pf3eqn3} implies that
\begin{eqnarray*}
\lefteqn{\|U_m(F)\|_p \le \sum_{j=m+1}^\infty \left\|\int_\XX \Phi_{2^{j}}(gb_{2^{j}};\circ,y)F(y)d\nu(y)\right\|_p}\\
&\le& c2^{-n\a/p'}\sum_{j=m+1}^\infty2^{-j(\beta-\a/p')}\|F\|_{\nu;\XX,p} \le c2^{-m\beta}2^{\a(m-n)/p'}\|F\|_{\nu;\XX,p}.
\end{eqnarray*}
\end{Proof}

We pause in our discussion to show that \eref{heatlowbd} implies a lower bound on the sum $\sum_{\ell_j\le L}\phi_j^2(x)$.
\begin{lemma}\label{tauberlemma}
Let $C>0$, $\{a_j\}$ be a sequence of nonnegative numbers such that $\sum_{j=0}^\infty \exp(-\ell_j^2t)a_j$ converges for $t\in (0,1]$. Then
\be\label{genchristlowbd}
c_1L^C\le \sum_{\ell_j\le L}a_j \le c_2L^C, \qquad L>0,
\ee
if and only if
\be\label{genheatlowbd}
c_3t^{-C/2}\le \sum_{j=0}^\infty \exp(-\ell_j^2t)a_j\le c_4t^{-C/2}, \qquad t\in (0,1].
\ee
In particular, \eref{heatlowbd} and \eref{singlegaussbd} imply that
\be\label{christfnbd}
c_1L^\a\le \sum_{\ell_j\le L}\phi_j^2(x)\le c_2L^\a, \qquad x\in\XX,\ L\ge 1.
\ee
\end{lemma}
\begin{Proof}\ %of Lemma~\ref{tauberlemma}
The fact that the upper bound in \eref{genheatlowbd} is equivalent to the upper bound in \eref{genchristlowbd} is proved in \cite[Proposition~4.1]{filbirbern}. 
In this proof only, let $s(u)=\sum_{\ell_j\le u} a_j$. Then
$$
\sum_{j=0}^\infty \exp(-\ell_j^2t)a_j=\int_0^\infty e^{-u^2t}ds(u).
$$
Since the sum converges, it is not difficult to verify by integration by parts that
\be\label{pf13eqn1}
\sum_{j=0}^\infty \exp(-\ell_j^2t)a_j=2t\int_0^\infty ue^{-u^2t}s(u)du.
\ee
If  \eref{genchristlowbd} holds, then $s(u)\ge cu^C$ for $u>0$, and
$$
2t\int_0^\infty ue^{-u^2t}s(u)du\ge 2ct\int_0^\infty u^{C+1}e^{-u^2t}du=ct^{-C/2}\int_0^\infty v^{C/2}e^{-v}dv =c_1t^{-C/2}.
$$
Thus, the lower bound in \eref{genchristlowbd} implies the lower bound in \eref{genheatlowbd}. 

In the remainder of this proof, it is convenient to let the constants retain their value, which might be different from what they were in the above part of the proof.  Let both the upper and lower inequalities in \eref{genheatlowbd} hold. Then the upper bound in \eref{genchristlowbd} holds also. We observe by integration by parts that for any $L>0$, $L^2t\ge C$, 
\begin{eqnarray*}
\int_L^\infty u^{C+1}e^{-u^2t}du&=&\frac{(L^2t)^{C/2}}{2t^{C/2+1}}\exp(-L^2t) +\frac{C}{2t}\int_L^\infty u^{C-1}e^{-u^2t}du\\
& \le& \frac{(L^2t)^{C/2}}{2t^{C/2+1}}\exp(-L^2t) +\frac{C}{2L^2t}\int_L^\infty u^{C+1}e^{-u^2t}du;
\end{eqnarray*}
i.e.,
$$
2t\int_L^\infty u^{C+1}e^{-u^2t}du\le \left(1-\frac{C}{2L^2t}\right)^{-1}(L^2t)^{C/2}\exp(-L^2t)t^{-C/2}.
$$
Thus, there exists $c_5$ such that 
$$
2t\int_L^\infty u^{C+1}e^{-u^2t}du\le \frac{c_3}{2c_2}t^{-C/2}, \qquad L^2t\ge c_5.
$$
We conclude from  the lower bound in \eref{genheatlowbd}, \eref{pf13eqn1}, and 
the upper bound in \eref{genchristlowbd}, that for $t, L>0$, $L^2t\ge c_5$,
\begin{eqnarray*}
c_3t^{-C/2}&\le& 2t\int_0^\infty ue^{-u^2t}s(u)du = 2t\int_0^L ue^{-u^2t}s(u)du+2t\int_L^\infty ue^{-u^2t}s(u)du\\
&\le& 2ts(L)\int_0^L ue^{-u^2t}du+2c_2t\int_L^\infty u^{C+1}e^{-u^2t}du\\
&\le& s(L)(1-\exp(-L^2t)) +c_3t^{-C/2}/2.
\end{eqnarray*}
Taking $t=c_5L^{-2}$, we obtain from here that $s(L)\ge c_6L^C$.
\end{Proof}

In the remainder of this paper, we adopt the following notation. Let $k^*\ge \max(2,(1/\a)\log_2(2c_2/c_1))$ be a fixed integer, where $c_1, c_2$ are the constants in \eref{christfnbd}. Then for $x\in\XX$,
$$
\sum_{\ell_j\le 2^{-k^*}L}\phi_j^2(x) \le c_22^{-\a k^*}L^\a \le (c_1/2)L^\a,
$$
and hence, \eref{heatlowbd} implies that
\be\label{levelchristbd}
\sum_{2^{-k^*}L\le \ell_j\le L}\phi_j^2(x)\ge (c_1/2)L^\a.
\ee
We further introduce $\tilde g(t):=h(t)-h(2^{k^*+1}t)$. Then $\tilde g(t)\ge 0$ for all $t\in\RR$, $\tilde g(t)=0$ if $0\le t\le 2^{-k^*-2}$ or $t\ge 1$, and $\tilde g(t)=1$ if $2^{-k^*-1}\le t\le 1/2$.
We note that 
\be\label{tildegtimesbineq}
\||\tilde g b_L\||_S\le cL^{-\beta}, \qquad L\ge 1.
\ee

The following lemma will be needed in the proof of Theorem~\ref{coefftheo}.

\begin{lemma}\label{gtildeloclemma}
Suppose that \eref{heatlowbd} holds. Let $\C\subset\XX$ be a finite set, $q=q(\C)\le 1$, and $\nu$ be a measure that associates the mass $q^\a$ with each $x\in\C$. Let \eref{ballmeasurecond}, \eref{heatkernint}, and \eref{singlegaussbd} hold.
Then $\nu\in {\cal M}_{q}$, and $\|\nu\|_{{\cal M}_q}\le c$, the constant being independent of $q$. Next, we assume in addition that \eref{heatlowbd} holds. Then for every integer $m$ with $2^m\ge q^{-1}$,
\be\label{awayest}
\sum_{y\in\C,\ x\not=y}|\Phi_{2^m}(\tilde g b_{2^m};x,y)| \le c(q2^{m})^{-S+\a}2^{m(\a-\beta)}, \qquad x\in\C,
\ee
and
\be\label{pf9eqn1}
\Phi_{2^m}(\tilde g b_{2^m};x,x)\ge c2^{m(\a-\beta)}, \qquad x\in\XX.
\ee
In particular, there exists $c_1>0$ such that for $2^mq\ge c_1$,
\be\label{diagdomfortildeg}
\sum_{y\in\C,\ x\not=y}|\Phi_{2^m}(\tilde g b_{2^m};x,y)|\le (1/2)|\Phi_{2^m}(\tilde g b_{2^m};x,x)|, \qquad x\in\C.
\ee
\end{lemma}
\begin{Proof}\ %of Lemma~\ref {gtildeloclemma}. 
If $x_0\in\XX$, $r>0$ and $B(x_0,r)\cap \C=\{y_1,\cdots,y_J\}$, then the balls $B(y_j, q/2)$ are disjoint, and $\cup_{j=1}^J B(y_j, q/2) \subset B(x_0,r+q/2)$. Using the fact that $\nu(B(x_0,r))=q^\a J$, and recalling \eref{ballmeasurelowbd}, we obtain
$$
\mu(B(x_0,r+q/2))\ge \mu(\cup_{j=1}^J B(y_j, q/2))=\sum_{j=1}^J\mu(B(y_j, q/2)) \ge cJq^\a =c\nu(B(x_0,r)).
$$
In turn, \eref{mzmeasequdef}  now implies that $\nu\in {\cal M}_{q}$, and $\|\nu\|_{{\cal M}_q}\le c$.
Since every point $y\in\C$ with $y\not=x$ is in $\Delta(x,q)$,  \eref{tildegtimesbineq} and \eref{phiintaway}, used with $q$ in place of $r$ and $d$, $2^m$ in place of $L$, imply that
$$
q^\a\sum_{y\in\C,\ x\not=y}|\Phi_{2^m}(\tilde g b_{2^m};x,y)| \le c(q2^m)^{-S+2\a}2^{-m\beta}= cq^\a (q2^m)^{-S+\a}2^{m(\a-\beta)}. 
$$
This proves \eref{awayest}.

We recall that $\tilde g(t)=1$ if $2^{-k^*-1}\le t\le 1/2$ and $b(\ell_j)\ge c\ell_j^{-\beta}$ for $\ell_j\ge c$. Consequently, \eref{levelchristbd} implies that for any $m\ge c$, and $x\in\XX$,
$$
\Phi_{2^m}(\tilde g b_{2^m};x,x) =\sum_{2^{m-k^*-2}\le \ell_j\le 2^m}\tilde g(\ell_j/2^m)b(\ell_j)\phi_j^2(x) \ge c2^{-m\beta}\sum_{2^{m-1-k^*}\le \ell_j\le 2^{m-1}} \phi_j^2(x)\ge c2^{m(\a-\beta)}.
$$
This proves \eref{pf9eqn1}. Recalling that $S>\a$, we may choose $m$ to make $2^mq$ large enough, yet $\sim 1$, so that \eref{awayest} and \eref{pf9eqn1} lead to  \eref{diagdomfortildeg}.
\end{Proof}

%5.2
\subsection{Diffusion polynomials}\label{diffpolysect}
In this section, we summarize various properties of the diffusion polynomials, and approximation by these.
The first statement is only a simple corollary of Theorem~\ref{kerntheo}.
\begin{cor}\label{mzcor}
Let $1\le p\le \infty$, $d>0$, $H$, and the other conditions be as in Theorem~\ref{kerntheo}, and $\nu\in {\cal M}_{d}$. Then for any $L>0$ and  $P\in \Pi_{L}$, 
\be\label{mutonusigmaopbd}
\|\sigma_L(H;\mu;f)\|_{\nu;\XX,p} \le c(1+(dL)^\a)^{1/p}\|\nu\|_{{\cal M}_{d}}^{1/p}\||H\||_S\|f\|_p,
\ee
\be\label{nutomusigmaopbd} 
\|\sigma_L(H;\nu;f)\|_p \le c(1+(dL)^\a)^{1/p'}\|\nu\|_{{\cal M}_{d}}^{1/p'}\||H\||_S\|f\|_{\nu;\XX,p}.
\ee
In particular, if $P\in \Pi_L$ then
\be\label{polymzineq}
\|P\|_{\nu;\XX,p}\le c(1+(dL)^\a)^{1/p}\|\nu\|_{{\cal M}_{d}}^{1/p}\|P\|_p.
\ee
\end{cor}
\begin{Proof}\ %of Corollary~\ref{mzcor}.
The estimates \eref{mutonusigmaopbd} and \eref{nutomusigmaopbd} follow from Lemma~\ref{rieszthorinlemma}, \eref{kernbdest}, and \eref{phiinttotal}. Let $P\in \Pi_L$. Then $\sigma_{2L}(h;\mu;P)=P$. We use \eref{mutonusigmaopbd} with $2L$ in place of $L$, $h$ in place of $H$, and $P$ in place of $f$ to deduce \eref{polymzineq}. 
\end{Proof}

The next lemma states some estimates for different pseudo--derivatives of diffusion polynomials. 
\begin{lemma}\label{gderlemma}
Let $\beta>\gamma\ge 0$, $L>0$, $P\in \Pi_L$, and \eref{ballmeasurecond},  \eref{singlegaussbd} hold.\\
{\rm (a)}  For any $r\ge 0$,
\be\label{pseudobern}
\|(\Delta^*)^rP\|_p \le cL^r\|P\|_p.
\ee
{\rm (b)} If $G$ is a kernel of type $\beta$, and ${\cal D}_G$ is the operator defined in Definition~\ref{gderdef}, then
\be\label{gderbd}
\|{\cal D}_GP\|_p \le cL^{\beta-\gamma}\|(\Delta^*)^\gamma P\|_p.
\ee
\end{lemma}
\begin{Proof}\ %of Lemma~\ref{gderlemma}
 Part (a) is proved in \cite{mauropap}. We will prove part (b). In this proof only, let $n\ge 1$ be an integer such that $L\le 2^{n-1}$. In this proof only, let $ b_\gamma(t)=(1+|t|)^\gamma b(t)$, $t\in\RR$. Then $b_\gamma^{-1}$ is a mask of type $\gamma-\beta<0$.
For $x\in\XX$, we have
\bea\label{pf11eqn2}
{\cal D}_GP(x)&=&\sum_{j=0}^{\infty} h\left(\frac{\ell_j}{2^n}\right)\frac{\ip{P}{\phi_j}}{b(\ell_j)}\phi_j(x)\nonumber\\
&=& \sum_{j=0}^\infty h\left(\frac{\ell_j}{2^n}\right)\frac{\ip{(\Delta^*)^\gamma P}{\phi_j}}{b(\ell_j)(1+\ell_j)^\gamma}\phi_j(x)\nonumber\\
&=&\int_\XX \Phi_{2^n}(h/b_{\gamma,2^n}; x,y)(\Delta^*)^\gamma P(y)d\mu(y).
\eea
We deduce \eref{gderbd} using  \eref{btimeshint} with $b_\gamma^{-1}$, $\gamma-\beta<0$ in place of $\beta$, and Lemma~\ref{rieszthorinlemma} with $\nu_1=\nu_2=\mu$. 

\end{Proof}
 
Even though a product of two diffusion polynomials is not necessarily a diffusion polynomial, the ``product assumption'' allows us to estimate the error in discretizing an integral of the product of such polynomials using a quadrature measure. This is summarized in the next lemma.

\begin{lemma}\label{quaderrlemma}
Let $L>0$, and \eref{ballmeasurecond},  \eref{singlegaussbd} hold. For any $p, r$, $1\le p\le r\le\infty$ and $P\in\Pi_L$,
\be\label{nikolskii}
\|P\|_r\le cL^{\a(1/p-1/r)}\|P\|_p.
\ee
We assume further that the product assumption holds. If $\nu$ is a quadrature measure of order $AL$, $|\nu|(\XX)\le c$, and $P_1,P_2\in\Pi_L$ then for any $p, r$, $1\le p, r\le\infty$ and any positive number $R>0$,
\be\label{polyquaderr}
\left|\int_\XX P_1P_2d\mu-\int_\XX P_1P_2d\nu\right| \le  c_1L^{2\a}\e_L\|P_1\|_p\|P_2\|_r \le c(R)L^{-R}\|P_1\|_p\|P_2\|_r.
\ee
\end{lemma}

\begin{Proof}\ %of Lemma~\ref{quaderrlemma}
Since 
$$
P(x)=\int_\XX P(y)\Phi_{2L}(x,y)d\mu(y),
$$
\eref{kernlocest} implies that $\|P\|_\infty\le cL^\a \|P\|_1$. Therefore, the convexity inequality (cf. \eref{convexityineq}) implies that $\|P\|_\infty \le cL^{\a/p}\|P\|_p$. If $r<\infty$, then
$$
\|P\|_r^r=\int_\XX |P(x)|^rd\mu(x)\le \|P\|_\infty^{r-p}\|P\|_p^p\le cL^{\a(r/p-1)}\|P\|_p^r.
$$
This proves \eref{nikolskii}. 

Next, we assume that the product assumption holds. Let $P_1=\sum_{\ell_m\le L}a_j\phi_j$, $P_2=\sum_{\ell_k\le L}d_k\phi_k$, and $Q_{j,k}\in \Pi_{AL}$ be found so that $\|\phi_j\phi_k-Q_{j,k}\|_\infty\le 2\dist(\infty;\phi_j\phi_k,\Pi_{AL})\le 2\e_L$.
Then, with $Q:=\sum_{j,k}a_jd_kQ_{j,k}$, we have for every $x\in\XX$,
\be\label{pf5eqn1}
|P_1(x)P_2(x)-Q(x)| =\left|\sum_{j,k}a_jd_k\left(\phi_j(x)\phi_k(x)-Q_{j,k}(x)\right)\right|\le 2\e_L\sum_{j,k}|a_j||d_k|.
\ee
In view of \eref{christupbd}, 
$$
|\{\ell_m\ :\ \ell_m\le L\}| =\sum_{\ell_m\le L}\int_\XX\phi_m^2(x)d\mu(x) \le cL^\a.
$$
Therefore, we conclude using \eref{nikolskii} and \eref{pf5eqn1} that 
$$
\|P_1P_2-Q\|_\infty\le 2\e_L\sum_{j,k}|a_j||d_k|\le cL^\a\e_L\|{\bf a}\|_{\ell^2}\|{\bf d}\|_{\ell^2}=cL^\a\e_L\|P_1\|_2\|P_2\|_2\le cL^{2\a}\e_L\|P_1\|_p\|P_2\|_r.
$$
Recalling that $|\nu|(\XX)\le c$, and $\int_\XX Qd\mu=\int_\XX Qd\nu$, we deduce that
\begin{eqnarray*}
\lefteqn{\left|\int_\XX P_1(x)P_2(x)d\mu(x)-\int_\XX P_1(x)P_2(x)d\nu(x)\right|}\\
&=&\left|\int_\XX (P_1(x)P_2(x)-Q(x))d\mu(x)-\int_\XX (P_1(x)P_2(x)-Q(x))d\nu(x)\right|\\
&\le&c\|P_1P_2-Q\|_\infty \le cL^{2\a}\e_L\|P_1\|_p\|P_2\|_r.
\end{eqnarray*}
The product assumption implies that $L^{2\a+R}\e_L\le c$, leading thereby to \eref{polyquaderr}.
\end{Proof}

Next, we prove a result regarding approximation by diffusion polynomials. Part (a) of this  result is essentially proved in \cite{mauropap}; we prove it again for the sake of completeness.

\begin{prop}\label{approxlemma}
For $1\le p\le\infty$, $f\in X^p$, $L>0$, $r>0$,  and \eref{ballmeasurecond},  \eref{singlegaussbd} hold. \\
{\rm (a)} We have
\be\label{jacksonest}
\|f-\sigma_L(f)\|_p +L^{-r}\|(\Delta^*)^r\sigma_L(f)\|_p \le c\omega_r(p;f,1/L).
\ee
In particular, if $f\in W^p_r$, then
\be\label{polyfavard}
\dist(p;f,\Pi_L)\le \|f-\sigma_L(f)\|_p \le cL^{-r}\|(\Delta^*)^rf\|_p.
\ee
{\rm (b)} If $f\in W^p_r$, $P\in \Pi_L$ satisfies $\|f-P\|_p\le \e$, then 
\be\label{freudest}
\|(\Delta^*)^r f- (\Delta^*)^rP\|_p\le c\{L^r\e+\dist(p;(\Delta^*)^rf, \Pi_{L/2})\}.
\ee
In particular, $\|(\Delta^*)^rP\|_p\le c(L^r\e+\|(\Delta^*)^rf\|_{p})$. \\
{\rm (c)} We assume in addition that the product assumption holds. Let $\nu$ be a $1/L$--regular quadrature measure of order $AL$.  For any $f\in W^\infty_r$,
\be\label{discapproxbd}
\|f-\sigma_L(\nu;f)\|_\infty\le cL^{-r}\{\|f\|_\infty+\|(\Delta^*)^rf\|_{\infty}\}.
\ee
If $f\in X^\infty$, then
\be\label{disckfuncreal}
\|f-\sigma_L(\nu;f)\|_\infty +L^{-r}\|(\Delta^*)^r\sigma_L(\nu;f)\|_\infty \le c\{\omega_r(\infty;f,L^{-1})+L^{-r}\|f\|_\infty\}.
\ee
\end{prop}
\begin{Proof}\ %of Proposition~\ref{approxlemma}
First, we prove \eref{polyfavard}. This proof is the same as that of \cite[(6.4)]{mauropap}. Thus, let $J$ be the greatest integer with $2^J\le L$. In this proof only, let $g_j(t)=g(t)/(2^{-j}+|t|)^r$, $t\in\RR$. Recalling that $g$ is supported on $[1/4,1]\cup [-1,-1/4]$, we see that $\||g_j\||_S\le c$. Hence, \eref{sigmaopbd} implies that
$$
\|\sigma_{2^{j}}(g; f)\|_p=2^{-jr}\|\sigma_{{2^j}}(g_j; (\Delta^*)^rf)\|_p \le c2^{-jr}\|(\Delta^*)^rf\|_p.
$$
Hence, 
\begin{eqnarray*}
\dist(p;f,\Pi_L)&\le& \dist(p;f,\Pi_{2^J}) \le \|f-\sigma_{2^J}(f)\|_p\le \sum_{j=J+1}^\infty \|\sigma_{2^{j}}(g; f)\|_p\\
&\le& c2^{-Jr}\|(\Delta^*)^rf\|_p\le cL^{-r}\|(\Delta^*)^rf\|_p.
\end{eqnarray*}
If $P\in\Pi_{L/2}$ is chosen so that $\|f-P\|_p\le 2\dist(p;f,\Pi_{L/2})$, then \eref{sigmaopbd} implies that
$$
\|f-\sigma_L(f)\|_p =\|f-P-\sigma_L(f-P)\|_p \le c\|f-P\|_p\le c\dist(p;f,\Pi_{L/2})\le cL^{-r}\|(\Delta^*)^rf\|_p.
$$
This proves \eref{polyfavard}.
In particular, we note that if $Q\in\Pi_{L/2}$ is chosen so that $\|(\Delta^*)^r(f-Q)\|_p\le 2\dist(p;(\Delta^*)^rf,\Pi_{L/2})$, then
\be\label{pf8eqn2}
\|f-\sigma_L(f)\|_p =\|f-Q-\sigma_L(f-Q)\|_p\le cL^{-r}\|(\Delta^*)^r(f-Q)\|_p\le cL^{-r}\dist(p;(\Delta^*)^rf,\Pi_{L/2}).
\ee
Next, let $f_1$ be chosen so that $\|f-f_1\|_p+L^{-r}\|(\Delta^*)^rf_1\|_p \le 2\omega_r(p;f,1/L)$. Then using \eref{sigmaopbd} and \eref{pseudobern}, we deduce that
\begin{eqnarray*}
\lefteqn{\|f-\sigma_L(f)\|_p +L^{-r}\|(\Delta^*)^r\sigma_L(f)\|_p}\\
& \le& \|f-f_1-\sigma_L(f-f_1)\|_p +\|f_1-\sigma_L(f_1)\|_p +L^{-r}\left(\|(\Delta^*)^r\sigma_L(f-f_1)\|_p +\|(\Delta^*)^r\sigma_L(f_1)\|_p\right)\\
&\le& c\{\|f-f_1\|_p +L^{-r}\|(\Delta^*)^rf_1\|_p + \|\sigma_L(f-f_1)\|_p +L^{-r}\|\sigma_L((\Delta^*)^rf_1)\|_p\}\\
&\le& c\{\|f-f_1\|_p +L^{-r}\|(\Delta^*)^rf_1\|_p\}\le c\omega_r(p;f,1/L).
\end{eqnarray*}
This proves \eref{jacksonest}.

Next, we prove part (b). In view of \eref{pseudobern}, \eref{sigmaopbd}, and \eref{pf8eqn2},
\begin{eqnarray*}
\|(\Delta^*)^rP-(\Delta^*)^r f\|_p &\le& \|(\Delta^*)^r(P-\sigma_{L}(f))\|_p +\|(\Delta^*)^r(f-\sigma_{L}(f))\|_p\\
&=&\|(\Delta^*)^r(P-\sigma_{L}(f))\|_p +\|(\Delta^*)^r(f)-\sigma_{L}((\Delta^*)^r(f))\|_p\\
&\le& cL^r\|P-\sigma_{L}(f)\|_p +c_1\dist(p;(\Delta^*)^rf,\Pi_{L/2})\\
&\le& cL^r\|P-f\|_p+cL^r\|f-\sigma_{L}(f)\|_p+c_1\dist(p;(\Delta^*)^rf,\Pi_{L/2})\\
&\le& cL^r\e + c_1\dist(p;(\Delta^*)^rf,\Pi_{L/2}).
\end{eqnarray*}
This proves part (b).

To prove part (c), let $P\in\Pi_{L/2}$ be arbitrary. Since
$$
P(x)=\int_\XX P(y)\Phi_L(x,y)d\mu(y), \qquad x\in\XX,
$$
we obtain from \eref{polyquaderr} (with $r$ in place of $R$) and \eref{kernbdest} that for every $x\in\XX$,
\bea\label{pf8eqn4}
|P(x)-\sigma_L(\nu;P,x)|&=&\left|\int_\XX P(y)\Phi_L(x,y)d\mu(y)-\int_\XX P(y)\Phi_L(x,y)d\nu(y)\right|\nonumber\\
&\le& c_1L^{-r}\|P\|_\infty\|\Phi_L(x,\circ)\|_1\le cL^{-r}\|P\|_\infty.
\eea
Hence, if $f\in W^\infty_r$,
\bea\label{pf8eqn1}
\|f-\sigma_L(\nu;f)\|_\infty&\!\!\le\!\!& \|f-\sigma_{L/2}(f)\|_\infty +\|\sigma_L(\nu;f-\sigma_{L/2}(f))\|_\infty\nonumber\\
&&\qquad + \|\sigma_{L/2}(f)-\sigma_L(\nu;\sigma_{L/2}(f))\|_\infty \nonumber\\
&\le& c\{\|f-\sigma_{L/2}(f)\|_\infty +L^{-r}\|\sigma_{L/2}(f)\|_\infty\}\nonumber\\
&\le& cL^{-r}\{\|(\Delta^*)^rf\|_\infty +\|f\|_\infty\}.
\eea
This proves \eref{discapproxbd}.
Next, let $f\in X^\infty$, and 
$$\|f-f_1\|_\infty +L^{-r}\|(\Delta^*)^rf_1\|_{\infty}\le 2\omega_r(\infty;f,1/L).$$
Then  using \eref{nutomusigmaopbd} and \eref{pf8eqn1} (with $f_1$ in place of $f$), we obtain
\bea\label{pf8eqn3}
\|f-\sigma_L(\nu;f)\|_\infty &\le& \|f-f_1\|_\infty +\|\sigma_L(\nu;f-f_1)\|_\infty +\|f_1-\sigma_L(\nu;f_1)\|_\infty\nonumber\\
& \le& c\{ \|f-f_1\|_\infty +L^{-r}\|(\Delta^*)^rf_1\|_{\infty}+L^{-r}\|f_1\|_\infty\}\nonumber\\
 &\le& c\{\omega_r(\infty;f,L^{-1})+L^{-r}\|f\|_\infty\}.
\eea
Applying \eref{pf8eqn1} with $f_1$ in place of $f$, and using part (b) of this proposition, we see that
\begin{eqnarray*}
\|(\Delta^*)^rf_1-(\Delta^*)^r\sigma_L(\nu;f_1)\|_\infty &\le& c\{\|(\Delta^*)^rf_1\|_\infty +\|f_1\|_\infty +\|(\Delta^*)^rf_1\|_\infty\}\\
&\le& c\{\|f-f_1\|_\infty +\|(\Delta^*)^rf_1\|_\infty +\|f\|_\infty\}.
\end{eqnarray*}
Hence, using \eref{pseudobern} and the uniform boundedness of the operators $\sigma_L(\nu)$, we obtain
\begin{eqnarray*}
\|(\Delta^*)^r\sigma_L(\nu;f)\|_\infty &\!\!\le\!\!&\|(\Delta^*)^r\sigma_L(\nu;f-f_1)\|_\infty +\|(\Delta^*)^rf_1-(\Delta^*)^r\sigma_L(\nu;f_1)\|_\infty+\|(\Delta^*)^rf_1\|_\infty \\
&\le& c\{L^r\|\sigma_L(\nu;f-f_1)\|_\infty +\|f-f_1\|_\infty +\|(\Delta^*)^rf_1\|_\infty +\|f\|_\infty\}\\
&\le& c\{L^r\|f-f_1\|_\infty +\|(\Delta^*)^rf_1\|_\infty +\|f\|_\infty\}\\
&\le& cL^r\{\omega_r(\infty;f,1/L) +L^{-r}\|f\|_\infty\}.
\end{eqnarray*}
The estimate \eref{disckfuncreal} follows from this estimate and \eref{pf8eqn3}.
\end{Proof}

%6
\bhag{Proofs of the main results}\label{proofsect}
In this section, we assume all the assumptions made in Section~\ref{manifoldsect}, namely, that \eref{ballmeasurecond}, \eref{heatkernint}, \eref{singlegaussbd}, \eref{heatgradest}, and the product assumption hold. We start with the proof of Theorem~\ref{directtheofirst}. Let ${\bf W}^*=\{w^*_y\}_{y\in\C^*}$, and $\nu^*$ be the measure that associates with each $y\in\C$ the mass $w^*_y$. As explained in Section~\ref{abstractsect}, the eignet $\GG(\C^*;{\bf W}^*; P)$ can be written more concisely as 
$$
\GG(\C^*; {\bf W}^*;P,x)=:\GG(\nu^*;P,x):=\GG(G;\nu^*;P,x)=\int_\XX ({\cal D}_GP)(y)G(x,y)d\nu^*(y), \qquad  x\in\XX.
$$
The condition that ${\bf W}^*$ is a $1/L$--regular set of quadrature weights of order $2AL$ corresponding to $\C^*$ can be stated more concisely in the form that $\nu^*\in {\cal M}_{1/L}$, $\|\nu^*\|_{{\cal M}_{1/L}}\le c$, and $\nu^*$ is a quadrature measure of order $2AL$.    Theorem~\ref{directtheofirst} then takes the form of the following Theorem~\ref{directtheo}. 

\begin{theorem}\label{directtheo}
Let $L>0$,  $\nu^*\in {\cal M}_{1/L}$, $\|\nu^*\|_{{\cal M}_{1/L}}\le c$, and $\nu^*$ be a quadrature measure of order $2AL$.   Let $1\le p\le\infty$, $\beta>\a/p'$, $0\le r <\beta$, $f\in W^p_r$. Let $P\in\Pi_L$ satisfy $\|f-P\|_p\le cL^{-r}\|(\Delta^*)^rf\|_p$. 
Then
\be\label{networkdirectbd}
\|f-\GG(\nu^*;P)\|_p\le cL^{-r}\|(\Delta^*)^rf\|_p.
\ee
\end{theorem}

The following lemma summarizes some of the major details of the proof of this theorem, so as to be applicable in the proof of some of the other results in Section~\ref{mainsect}.

\begin{lemma}\label{polytoeignetlemma}
Let $n\ge 1$ be an integer,  $\nu\in {\cal M}_{2^{-n}}$, $\|\nu\|_{{\cal M}_{2^{-n}}}\le c$.    Let $1\le p\le\infty$, $\beta>\a/p'$, $0\le r <\beta$, $P\in\Pi_{2^n}$. We have
\be\label{polyeignetdiff}
\left\|\int_\XX \{G(x,y)-\Phi_{2^n}(hb_{2^n};x,y)\}{\cal D}_GP(y)d\nu(y)\right\|_p\le c2^{-nr}\|(\Delta^*)^rP\|_p\le c\|P\|_p.
\ee
In addition, if $\nu$ is a quadrature measure of order $A2^n$, and $R>0$, then
\bea\label{nutomudiff}
\lefteqn{\left|\int_\XX \Phi_{2^{n}}(hb_{2^{n}};x,y){\cal D}_GP(y)d\nu(y)-\int_\XX \Phi_{2^{n}}(hb_{2^{n}};x,y){\cal D}_GP(y)d\mu(y)\right|}\nonumber\\
&\le& c(R)2^{-n(R+r)}\|(\Delta^*)^rP\|_p\le c(R)2^{-nR}\|P\|_p,
\eea
and
\be\label{peignetapprox}
\|P-\GG(\nu;P)\|_p\le c2^{-nr}\|(\Delta^*)^rP\|_p\le c\|P\|_p.
\ee
If $0<\gamma<\beta-\a/p'$, and $\gamma\le r\le \beta$, then
\be\label{peignetderapprox}
\|(\Delta^*)^\gamma P-(\Delta^*)^\gamma\GG(\nu;P)\|_p\le c2^{-n(r-\gamma)}\|(\Delta^*)^rP\|_p.
\ee
\end{lemma}

\begin{Proof}\ %of Lemma~\ref{polytoeignetlemma}
Since ${\cal D}_GP\in\Pi_{2^n}$, we conclude using \eref{polymzineq}, \eref{gderbd}, and \eref{pseudobern} with $2^{-n}$ in place of $d$, $2^n$ in place of $L$ and $r$ in place of $\gamma$ that 
$$
\|{\cal D}_GP\|_{\nu;\XX,p}\le c\|{\cal D}_GP\|_p\le c2^{n(\beta-r)}\|(\Delta^*)^rP\|_p \le c2^{n\beta}\|P\|_p.
$$
The estimate \eref{polyeignetdiff} follows from this and Proposition~\ref{networkkernprop}(b), used with $m=n$, ${\cal D}_GP$ in place of $F$. 

Next, for each $x\in\XX$, \eref{polyquaderr} (with $R+\beta$ in place of $R$) and the last estimate in \eref{btimeshint}  imply that
\begin{eqnarray*}
\lefteqn{\left|\int_\XX \Phi_{2^{n}}(hb_{2^{n}};x,y){\cal D}_GP(y)d\nu(y)-\int_\XX \Phi_{2^{n}}(hb_{2^{n}};x,y){\cal D}_GP(y)d\mu(y)\right|}\\
&\le& c(R)2^{-n(R+\beta)}\|\Phi_{2^{n}}(hb_{2^{n}};x,\circ)\|_{1}\|{\cal D}_GP\|_p\le c_1(R)2^{-n(R+r)}\|(\Delta^*)^rP\|_p. 
\end{eqnarray*}
This proves the first inequality in \eref{nutomudiff}; the second follows from \eref{pseudobern}.

In this proof only, we write $\tilde\nu=\mu-\nu$, and observe that $\|\tilde\nu\|_{{\cal M}_{2^{-n}}}\le c$.
In view of  \eref{polyaseignet}, we obtain
\bea\label{pf6eqn1}
\lefteqn{P(x)-\GG(\nu;P,x)=\int_\XX G(x,y){\cal D}_GP(y)d\tilde\nu(y)}\nonumber\\
&=&\int_\XX \{G(x,y)-\Phi_{2^n}(hb_{2^n};x,y)\}{\cal D}_GP(y)d\tilde\nu(y) + \int_\XX \Phi_{2^n}(hb_{2^n};x,y){\cal D}_GP(y)d\tilde\nu(y).
\eea
Using the first estimate in \eref{polyeignetdiff} with $\tilde\nu$ in place of $\nu$,  we obtain
\be\label{pf6eqn2}
\left\|\int_\XX \{G(x,y)-\Phi_{2^n}(hb_{2^n};x,y)\}{\cal D}_GP(y)d\tilde\nu(y)\right\|_p\le c2^{-nr}\|(\Delta^*)^rP\|_p.
\ee
Together with \eref{nutomudiff}, \eref{pf6eqn1}, this implies \eref{peignetapprox}.

In the remainder of this proof only, let $G_\gamma(x,y)$ be defined formally by \\
$G_\gamma(x,y)=\sum_{j=0}^\infty (1+\ell_j)^\gamma b(\ell_j)\phi_j(x)\phi_j(y)$. Then $G_\gamma$ is clearly a kernel of type $\beta-\gamma>\a/p'$. Let $P\in\Pi_\infty$. For $y\in\XX$, we have
$$
{\cal D}_GP(y) =\sum_{j=0}^\infty \frac{\ip{P}{\phi_j}}{b(\ell_j)}\phi_j(y) = \sum_{j=0}^\infty \frac{\ip{P}{\phi_j}(1+\ell_j)^\gamma}{(1+\ell_j)^\gamma b(\ell_j)}\phi_j(y)={\cal D}_{G_\gamma}((\Delta^*)^\gamma P)(y).
$$
Consequently,  we obtain for $x\in\XX$,
\begin{eqnarray*}
(\Delta^*)^\gamma\GG(G;\nu;P,x) &=&\int_\XX G_\gamma(x,y){\cal D}_GP(y)d\nu(y) =\int_\XX G_\gamma(x,y){\cal D}_{G_\gamma}((\Delta^*)^r P)(y)d\nu(y)\\
&=&\GG(G_\gamma;\nu;(\Delta^*)^\gamma P).
\end{eqnarray*}
The estimate \eref{peignetderapprox} now follows easily from \eref{peignetapprox}, used with $(\Delta^*)^\gamma P$ in place of $P$, $r-\gamma$ in place of $r$.
\end{Proof}

\noindent
\textsc{Proof of Theorem~\ref{directtheo} (and hence, Theorem~\ref{directtheofirst}).} 
In this proof only, let $n$ be the greatest integer such that $2^n\le L$. Then $\nu^*$ is also a $2^{-n}$-- regular quadrature measure of order $2A2^n$, and $\|\nu^*\|_{{\cal M}_{2^{-n}}}\le c$. In view of Proposition~\ref{approxlemma}(b), $\|(\Delta^*)^rP\|_p \le c\|(\Delta^*)^rf\|_p$. 
Our choice of $P$ and \eref{peignetapprox} now imply \eref{networkdirectbd}.
\qed

\noindent
\textsc{Proof of Theorem~\ref{jacksontheofirst}.} We note that in our current notation, $\GG_L(f)=\GG(\nu^*;\sigma_L(f))$. We let $n$ be as in the proof of Theorem~\ref{directtheo}. Hence, using \eref{peignetapprox} and Proposition~\ref{approxlemma}(a), we obtain
\bea\label{pf6eqn3}
\|f-\GG(\nu^*;\sigma_L(f))\|_p&\le& \|f-\sigma_L(f)\|_p +\|\sigma_L(f)-\GG(\nu^*;\sigma_L(f))\|_p\nonumber\\
&\le& \|f-\sigma_L(f)\|_p + cL^{-r}\|(\Delta^*)^r\sigma_L(f)\|_p \le c\omega_r(p;f, 1/L).
\eea
Since it is obvious that $\omega_r(p;f, 1/L)\le \|f\|_p$ (by choosing $f_1=0$ in the definition of $\omega_r)$, this implies also that $\|\GG(\nu^*;\sigma_L(f))\|_p\le c\|f\|_p$.

Using \eref{peignetderapprox} with $r=\gamma$ and $\sigma_L(f)$ in place of $P$, we obtain
$$
\|(\Delta^*)^r\GG(\nu^*;\sigma_L(f))-(\Delta^*)^r\sigma_L(f)\|_p  \le c\|(\Delta^*)^r\sigma_L(f)\|_p.
$$
Hence, using Proposition~\ref{approxlemma}(a) again,
$$
\|(\Delta^*)^r\GG(\nu^*;\sigma_L(f))\|_p \le c\|(\Delta^*)^r\sigma_L(f)\|_p\le cL^{r}\omega_r(p;f,1/L).
$$
Together with \eref{pf6eqn3}, this implies \eref{jacksonestfirst}.

Next, we turn to part (b). In this part of the proof, let $\nu$ be the measure that associates the mass $w_y$ with each $y\in\C$, so that $\|\nu\|_{{\cal M}_{1/L}}\le c$. Then in our current notation, $$\tilde\GG_L(\C;{\bf W};f)=\GG(\C^*; {\bf W}^*; \sigma_L(\C;  {\bf W};f))=\GG(\nu^*;\sigma_L(\nu;f)).$$ Using \eref{peignetapprox}, \eref{nutomusigmaopbd} with 
$d=1/L$,  $H=h$, we obtain
$$
\|\GG(\nu^*;\sigma_L(\nu;f))\|_p \le c\|\sigma_L(\nu;f)\|_p \le c\|f\|_{\nu;\XX,p}.
$$
This proves \eref{discstability}. The proof of \eref{discjacksonestfirst} is the same as that of \eref{jacksonestfirst}, except that we have to use Proposition~\ref{approxlemma}(c) instead, and the estimates are accordingly as claimed.
\qed

During the rest of this section, we assume that \eref{heatlowbd} (and hence, by Lemma~\ref{tauberlemma}, \eref{christfnbd}) holds.
Next, we prove Theorem~\ref{coefftheo}. This will be done using Lemma~\ref {gtildeloclemma} and the following general statement about the inverse of matrices. Proposition~\ref{matrixinvprop} is most probably not new, but we find it easier to prove it than to find a reference for it.

\begin{prop}\label{matrixinvprop}
Let $M\ge1$ be an integer, ${\bf A}$ be an $M\times M$ matrix whose $(i,j)$--th entry is $A_{i,j}$. $1\le p\le\infty$, and $\gamma\in [0,1)$. If
\be\label{diagdom}
\sum_{i=1\atop i\not=j}^M |A_{j,i}| \le \gamma |A_{j,j}|,\ \sum_{i=1\atop i\not=j}^M |A_{i,j}| \le \gamma|A_{j,j}|,  \qquad j=1,\cdots,M,
\ee
and $\lambda=\min_{1\le i\le M}|A_{i,i}|>0$, then ${\bf A}$ is invertible, and
\be\label{diagmatrixinvnorm}
\|{\bf A}^{-1}{\bf y}\|_{\ell^p}\le ((1-\gamma)\lambda)^{-1}\|{\bf y}\|_{\ell^p}, \qquad {\bf y}\in\RR^M.
\ee
\end{prop}
\begin{Proof}\ %of Proposition~\ref{matrixinvprop}.
Let ${\bf a}=(a_1,\cdots,a_M)\in\RR^M$, and $\y={\bf A}{\bf a}$. First, we consider the case $p=\infty$. Let $j^*$ be the index such that $|a_{j^*}|=\|{\bf a}\|_{\ell^\infty}$. Then, in view of the first estimate in \eref{diagdom}, we have
\begin{eqnarray*}
\|\y\|_{\ell^\infty}&\ge& |y_{j^*}| =\left|\sum_{i=1}^M A_{j^*,i}a_i\right|\ge
 |A_{j^*,j^*}||a_{j^*}| -\sum_{i=1\atop i\not=j^*}^M |A_{j^*,i}||a_i|\\ &\ge&  |A_{j^*,j^*}|(1-\gamma)\|{\bf a}\|_{\ell^\infty}\ge (1-\gamma)\lambda \|{\bf a}\|_{\ell^\infty}.
\end{eqnarray*}
Therefore, ${\bf A}$ is invertible. For every $\y$, there exists ${\bf a}={\bf A}^{-1}\y$. Applying the above chain of inequalities with this ${\bf a}$,  we have proved \eref{diagmatrixinvnorm} in the case $p=\infty$.

Next, using the second estimate in \eref{diagdom}, we obtain 
\begin{eqnarray*}
\|\y\|_{\ell^1}&=&\sum_{i=1}^M|y_i| =\sum_{i=1}^M\left|\sum_{j=1}^M A_{i,j}a_j\right|\\
&\ge& \sum_{j=1}^M |A_{j,j}||a_j| -\sum_{j=1}^M \sum_{i=1\atop i\not=j}^M |A_{i,j}||a_j|\\
&\ge& \sum_{j=1}^M |A_{j,j}|(1-\gamma)|a_j| \ge \lambda (1-\gamma)\|{\bf a}\|_{\ell^1}.
\end{eqnarray*}
This proves \eref{diagmatrixinvnorm} in the case $p=1$.

The intermediate cases, $1<p<\infty$, of \eref{diagmatrixinvnorm} follow from the Riesz--Thorin interpolation theorem.
\end{Proof}

\noindent
\textsc{Proof of Theorem~\ref{coefftheo}.}
In this proof only, let $\Psi=\sum_{y\in\C} a_yG(\circ,y)$, and $m$ be chosen so that $2^m\ge c_1q^{-1}$ and \eref{diagdomfortildeg} holds. Then,with $\tilde g$ as defined just before Lemma~\ref{gtildeloclemma},
$$
\Phi_{2^m}(\tilde g ;\Psi,x)=\sum_{j=0}^\infty \tilde g(\ell_j/2^m)\sum_{y\in\C} a_y b(\ell_j)\phi_j(y)\phi_j(x)=\sum_{y\in\C}a_y\Phi_{2^m}(\tilde g b_{2^m};x,y).
$$
In this proof only, let ${\bf d}$ denote the vector $(\Phi_{2^m}(\tilde g ;\Psi,x))_{x\in\C}$, where all vectors are treated as column vectors, and ${\bf A}$ denote the $|\C|\times|\C|$ matrix whose $(x,y)$-th entry is given by $\Phi_{2^m}(\tilde g b_{2^m};x,y)$. 
Then \eref{diagdomfortildeg} implies that \eref{diagdom} is satisfied with $\gamma=1/2$. Also, \eref{pf9eqn1} implies that $\min_{x\in\C}A_{x,x}\ge c2^{m(\a-\beta)}$, $x\in\C$. Therefore, Proposition~\ref{matrixinvprop} shows that ${\bf A}$ is invertible. Further, \eref{diagmatrixinvnorm} implies that
$$
\|{\bf a}\|_{\ell^p} \le  c2^{m(\beta-\a)}\|{\bf d}\|_{\ell^p}.
$$
Now, let $\nu$ be the measure as in Lemma~\ref{gtildeloclemma}. Then $\nu\in{\cal M}_q$. So, \eref{polymzineq} shows that for $2^m\ge c_1/q$,
$$
\|{\bf d}\|_{\ell^p} =q^{-\a/p}\|\Phi_{2^m}(\tilde g ;\Psi)\|_{\nu;\XX,p}\le cq^{-\a/p}(2^{m}q)^{\a/p}\|\Phi_{2^m}(\tilde g ;\Psi)\|_{p}.
$$
In view of \eref{sigmaopbd} applied with $\tilde g$ in place of $H$, $\|\Phi_{2^m}(\tilde g ;\Psi)\|_{p}\le c\|\Psi\|_p$. Hence, for $2^m\ge c_1/q$,
$$
\|{\bf a}\|_{\ell^p}\le c2^{m(\beta-\a/p')}\|\Psi\|_p.
$$
We may now choose $m$ with $2^m\sim q^{-1}$ to arrive at  \eref{coeffineq}.
\qed 

Next, we turn our attention to the proof of Theorem~\ref{equivtheo}. Towards this end, we recall the following theorem (\cite[Chapter~7, Theorem~9.1, also Chapter 6.7]{devlorbk}). Our assumption about the centers $\C_m$ in the definition of the spaces $\V_m$ being nested implies that the sequence of spaces  $\{\V_m\}$ satisfies the conditions listed in \cite[Chapter~7, (5.2)]{devlorbk} with the class $X^p$ in place of $X$ in \cite{devlorbk}, where the density assumption can be verified easily using \eref{polyaseignet} and the fact that $\delta(\C_m)\to 0$ as $m\to\infty$.  The statement of \cite[Chapter~7, Theorem~9.1]{devlorbk} is in terms of the Besov spaces in general, we apply it with the parameter $q=\infty$ there.

\begin{theorem}\label{devlortheo}
 Let $1\le p\le \infty$, $r>0$. Suppose that for some  $r>0$,
\be\label{favardest}
\dist (F, \V_m)\le cm^{-r}\|(\Delta^*)^rF\|_p, \qquad m=1,2,\cdots, \ F\in W^p_r,
\ee
and
\be\label{bernineq}
\|(\Delta^*)^r \Psi\|_p \le cm^{r}\|\Psi\|_p, \qquad \Psi\in\V_m,\ m=1,2,\cdots.
\ee
Then for $0<\gamma<r$, $F\in H^p_\gamma$ if and only if $\sup_{m\ge 1}m^{\gamma}\dist(F,\V_m)\le c(F)$.
\end{theorem}

Theorem~\ref{directtheo} (used with $\C_m$, ${\bf W}_m$ in place of $\C^*$, ${\bf W}^*$ respectively) already shows that \eref{favardest} holds. Thus, to complete the proof of Theorem~\ref{equivtheo}, we need to establish

\begin{theorem}\label{gberntheo}
Let $1\le p\le\infty$,  $0<r<\beta-\a/p'$, $\C\subset\XX$ be a finite set,  $q=q(\C)$, and $\{a_y\}_{y\in\C}\subset\RR$. Then
\be\label{gbernineq}
\|(\Delta^*)^r\sum_{y\in\C}a_yG(\circ,y)\|_p \le cq^{-r}\|\sum_{y\in\C}a_yG(\circ,y)\|_p.
\ee 
\end{theorem}

\begin{Proof}\ %of Theorem~\ref{gberntheo}
 Let $\nu\in{\cal M}_q$ be the measure as in  Lemma~\ref{gtildeloclemma}. In this proof only, let $\Psi=\sum_{y\in\C}a_yG(\circ,y)$. Then Proposition~\ref{networkkernprop} (b), used with  $n=\lfloor \log_2 (1/q)\rfloor$, shows that for any $F :\C\to\RR$,
$$
\left\|\int_{y\in\XX} \{G(\circ,y)-\Phi_{2^{m}}(hb_{2^{m}};\circ,y)\}F(y)d\nu(y)\right\|_p\le c2^{-m\beta}.2^{\a(m-n)/p'}\|F\|_{\nu;\XX,p}.
$$
Using $2^{-n}\sim q$, and the function $F$ defined by $F(y)=a_y$, $y\in\C$, this translates into
$$
\left\|q^\a\Psi-q^\a\sum_{y\in \C}a_y\Phi_{2^{m}}(hb_{2^{m}};\circ,y)\right\|_p\le c2^{-m\beta}(q2^m)^{\a/p'}q^{\a/p}\|{\bf a}\|_{\ell^p};
$$
i.e.,
$$
\left\|\Psi-\sum_{y\in \C}a_y\Phi_{2^{m}}(hb_{2^{m}};\circ,y)\right\|_p\le c2^{-m(\beta-\a/p')}\|{\bf a}\|_{\ell^p}.
$$
In view of \eref{coeffineq}, this yields
\be\label{pf10eqn1}
\left\|\Psi-\sum_{y\in \C}a_y\Phi_{2^{m}}(hb_{2^{m}};\circ,y)\right\|_p\le c2^{-m(\beta-\a/p')}q^{\a/p'-\beta}\|\Psi\|_p.
\ee

Next,  we note that the function $b_r(t):=(1+|t|)^rb(t)$, $t\in\RR$, is a mask of type $\beta-r$, and also that $(\Delta^*)^rG(\circ,y)=G(b_r;\circ,y)$, $y\in\XX$. Similarly, $(\Delta^*)^r\Phi_{2^{m}}(hb_{2^{m}};\circ,y)=\Phi_{2^{m}}(hb_{r,2^{m}};\circ,y)$. Hence, we may apply \eref{pf10eqn1} with $(\Delta^*)^r G(\circ,y)$ in place of $G$, $\beta-r$ in place of $\beta$, and deduce that
\be\label{pf10eqn2}
\left\|(\Delta^*)^r\Psi-(\Delta^*)^r\sum_{y\in \C}a_y\Phi_{2^{m}}(hb_{2^{m}};\circ,y)\right\|_p\le c2^{-m(\beta-r-\a/p')}q^{\a/p'-\beta+r}\|(\Delta^*)^r\Psi\|_p.
\ee
We now choose $m$ sufficiently large, so that $2^m\sim 1/q$,  and $c2^{-m(\beta-r-\a/p')}q^{\a/p'-\beta+r}\le 1/2$. Then \eref{pf10eqn1}, \eref{pf10eqn2} become
$$
\left\|\Psi-\sum_{y\in \C}a_y\Phi_{2^{m}}(hb_{2^{m}};\circ,y)\right\|_p\le c\|\Psi\|_p,
$$
and
$$
\left\|(\Delta^*)^r\Psi-(\Delta^*)^r\sum_{y\in \C}a_y\Phi_{2^{m}}(hb_{2^{m}};\circ,y)\right\|_p\le (1/2)\|(\Delta^*)^r\Psi\|_p.
$$
Since $\sum_{y\in \C}a_y\Phi_{2^{m}}(hb_{2^{m}};\circ,y)\in\Pi_{2^m}$, these estimates and \eref{pseudobern} lead to
$$
\|(\Delta^*)^r\Psi\|_p\le 2\left\|(\Delta^*)^r\sum_{y\in \C}a_y\Phi_{2^{m}}(hb_{2^{m}};\circ,y)\right\|_p\le c2^{mr}\left\|\sum_{y\in \C}a_y\Phi_{2^{m}}(hb_{2^{m}};\circ,y)\right\|_p\le c2^{mr}\|\Psi\|_p.
$$
Since $2^m\sim 1/q$, this implies \eref{gbernineq}.
\end{Proof}

\noindent
\textsc{Proof of Theorem~\ref{equivtheo}.} We note that Theorem~\ref{devlortheo}  is applicable in view of Theorem~\ref{directtheo} and Theorem~\ref{gberntheo}. The equivalence (a)$\Leftrightarrow$(c) follows from Theorem~\ref{devlortheo}. The implication (a)$\Rightarrow$(b) follows from Theorem~\ref{jacksontheofirst}. The implication (b)$\Rightarrow$(c) is clear. 

  In the case when $p=\infty$, the implication (d)$\Rightarrow$(c) is clear. The implication  (a)$\Rightarrow$(d) follows from Theorem~\ref{jacksontheofirst}.
\qed

\noindent
\textsc{Proof of Theorem~\ref{simapproxtheo}.} 
 Using \eref{peignetderapprox}, Theorem~\ref{gberntheo} (used with $\gamma$ in place of $r$), and Theorem~\ref{directtheo}, we obtain
\begin{eqnarray*}
\|(\Delta^*)^\gamma \sigma_m(f) -(\Delta^*)^\gamma \Psi_m\|_p&\le&\|(\Delta^*)^\gamma \sigma_m(f)- (\Delta^*)^\gamma \GG_m(f)\|_p +\|(\Delta^*)^\gamma \GG_m(f)-(\Delta^*)^\gamma \Psi_m\|_p\\
&\le&c\left\{m^{\gamma-r}\|(\Delta^*)^r\sigma_m(f)\|_{p} + m^\gamma\|\GG_m(f)-\Psi_m\|_p\right\}\\
&\le& c\left\{m^{\gamma-r}\|(\Delta^*)^r f\|_{p} + m^\gamma\|f-\GG_m(f)\|_p+m^\gamma\|f-\Psi_m\|_p\right\}\\
&\le&cm^{\gamma-r}\|(\Delta^*)^r f\|_{p}.
\end{eqnarray*}
In view of Proposition~\ref{approxlemma}, this leads to the desired estimate.
\qed

We end this section with the postponed proof of Proposition~\ref{datasetprop}.\\

\noindent\textsc{Proof of Proposition~\ref{datasetprop}.} In order to prove part (a), let (in this proof only)  $\C=\{x_k\}_{k=1}^M$. We define $\C_1^*=\C\cap \Delta(x_1,\e)$. By relabeling the set if necessary, we choose $x_2\in\C_1^*$, and set $\C_2^*=\C_1^*\cap\Delta(x_2,\e)$. Necessarily, $\rho(x_1,\C_2^*)\ge \e$ and $\rho(x_1,x_2)\ge \e$. Since $\C$ is finite, we may continue in this way at most $M$ times to obtain a subset $\tilde \C$ of $\C$ such that $q(\tilde\C)\ge \e$, and moreover, for any $x\in \C$, there is $y\in\tilde\C$ with $\rho(x,y)\le \e$; i.e., $\delta(\tilde\C, \C)\le \e$. It follows that 
$$
\delta(\C) \le \delta(\tilde\C)\le \delta(\C)+\e.
$$
This completes the proof of part (a). 

To prove part (b), we will use some notation which will be different from the rest of the proof. In view of the fact that $\delta(\C_1)\le (1/2)\delta(\C_0)\le q(\C_0)$, the points of $\C_0$ are  already at least $\delta(\C_1)$ separated from each other. Let $\C_1^{\#}$ be the subset of $\C_1\setminus \C_0$ comprising points which are at least $\delta(\C_1)$ away from any point in $\C_0$. Let $\C_1^+\subseteq \C_1^{\#}$ be selected as in part (a), so that 
\be\label{pf2eqn1}
\delta(\C_1^+,\C_1^{\#})\le \delta(\C_1)\le q(\C_1^+),
\ee
 and $\C_1^*:=\C_1^+\cup \C_0$.  Clearly, $\C_1^*\supseteq \C_0$, and $q(\C_1^*)\ge \delta(\C_1)$.  If $x\in \C_1$ and there is no point of $\C_0$ within $\delta(\C_1)$ of $x$, then $x\in \C_1^{\#}$. In view of \eref{pf2eqn1}, there is a point in $\C_1^+$ within $\delta(\C_1)$ of $x$. So, in any case, for any $x\in \C_1$, there is a point in $C_1^*$ within $\delta(\C_1)$ of $x$. Therefore, 
$$
\delta(\C_1)\le \delta(\C_1^*) \le 2\delta(\C_1) \le  2q(\C_1^*).
$$
This completes the proof of part (b).

To prove part (c), we note that there exist integers $\ell, n\ge 0$ such that
\be\label{pf2eqn2}
(2^\ell k)^{-1}\le \delta(\C_k)\le (2^{-n}k)^{-1}, \qquad k=1,2,\cdots.
\ee
In this proof only, we define $\C_k'=\C_{2^{k(\ell+n+1)}}$, $k=0,1,2,\cdots$.  Then it is clear that $\C_k'\subseteq \C_{k+1}'$ and it is easy to check using \eref{pf2eqn2} that $\delta(\C_{k+1}')\le (1/2)\delta(\C_k')$. With the construction as in the proof of part (a), we choose $C_0''\subseteq \C_0'$ such that 
$$
\delta(\C_1')\le (1/2)\delta(\C_0')\le (1/2)\delta(\C_0'')\le q(\C_0'').
$$
We then use part (b) with $\C_0''$ in place of $\C_0$ of part (b) and $\C_1'$ in place of $\C_1$ of part (b) to obtain $\C_1''\subset \C_1'$ such that $\C_1''\supseteq \C_0''$, $\delta(C_1')\le \delta(\C_2'')\le 2\delta(\C_1')\le 2q(\C_1'')$, and $\delta(\C_2')\le (1/2)\delta(\C_1')\le (1/2)\delta(\C_1'')$. Proceeding by induction, we construct an increasingly nested sequence $\{\C_k''\subseteq \C_k'\}$ with $\delta(\C_k'')\le 2\delta(\C_k')\le 2q(\C_k'')$. We observe that
\be\label{pf2eqn3}
(2^{k(\ell+n+1)+\ell})^{-1}\le \delta(\C_k')\le \delta(\C_k'') \le 2\delta(\C_k')\le 2(2^{k(\ell+n+1)+n})^{-1}.
\ee
If $m\ge 1$ is any integer, we find integer $k$ such that $2^{k(\ell+n+1)}\le m < 2^{(k+1)(\ell+n+1)}$, and define $\tilde\C_m =\C_k''$. Then $\C_m\supseteq \C_{2^{k(\ell+n+1)}}=\C_k'\supseteq \C_k''\supseteq \tilde\C_m$. Moreover, since the value of $k$ corresponding to $m$ does not exceed that corresponding to $m+1$, and the sequence $\{C_k''\}$ is increasingly nested, then $\tilde\C_m\subseteq\tilde\C_{m+1}$. It is easy to verify  from \eref{pf2eqn3} that $\delta(\tilde\C_m)\le 2\delta(\tilde\C_m)$ and that $\delta(\tilde\C_m)\sim 1/m$. 
\qed

%R

\end{document}